\documentclass{article} %
\usepackage[preprint]{colm2025_conference}
\usepackage{microtype}
\usepackage{hyperref}
\usepackage{url}
\usepackage{booktabs}
\usepackage{xspace}
\usepackage{multirow}
\usepackage{graphicx}
\usepackage{lineno}
\usepackage{amsmath}
\usepackage[T1]{fontenc}
\usepackage{subcaption}

\definecolor{darkblue}{rgb}{0, 0, 0.5}
\hypersetup{colorlinks=true, citecolor=darkblue, linkcolor=darkblue, urlcolor=darkblue}

\title{Improving In-Context Learning with Reasoning Distillation}

\author{Nafis Sadeq, Xin Xu, Zhouhang Xie \& Julian McAuley\\
UC San Diego \\
La Jolla, CA 92093, USA \\
\texttt{\{nsadeq,xinxucs,zhx022,jmcauley\}@ucsd.edu} \\
\AND
Byungkyu Kang, Prarit Lamba \& Xiang Gao \\
Intuit \\
Mountain View, CA 94043, USA\\
\texttt{\{Jay\_Kang,Prarit\_Lamba,Xiang\_Gao\}@intuit.com} \\
}

\newcommand{\proposed}{ReDis\xspace}

\begin{document}

\ifcolmsubmission
\linenumbers
\fi

\maketitle

\begin{abstract}
Language models rely on semantic priors to perform in-context learning, which leads to poor performance on tasks involving inductive reasoning. Instruction-tuning methods based on imitation learning can superficially enhance the in-context learning performance of language models, but they often fail to improve the model's understanding of the underlying rules that connect inputs and outputs in few-shot demonstrations. We propose \proposed, a reasoning distillation technique designed to improve the inductive reasoning capabilities of language models. Through a careful combination of data augmentation, filtering, supervised fine-tuning, and alignment, \proposed achieves significant performance improvements across a diverse range of tasks, including 1D-ARC, List Function, ACRE, and MiniSCAN. Experiments on three language model backbones show that \proposed outperforms equivalent few-shot prompting baselines across all tasks and even surpasses the teacher model, GPT-4o, in some cases. \proposed, based on the LLaMA-3 backbone, achieves relative improvements of 23.2\%, 2.8\%, and 66.6\% over GPT-4o on 1D-ARC, ACRE, and MiniSCAN, respectively, within a similar hypothesis search space. The code, dataset, and model checkpoints will be made available at \url{https://github.com/NafisSadeq/reasoning-distillation.git}.
\end{abstract}
\section{Introduction}
\label{sec:intro}
Scaling up transformer-based language models has unlocked the in-context learning (ICL) paradigm, where the model can learn to perform a previously unseen task from a few demonstrated examples~\citep{brown2020language,chowdhery2023palm,ouyang2022training}.
However, improving ICL capabilities remains a challenging task~\citep{gendron2024largelanguagemodelsstrong,li2025mirageevaluatingexplaininginductive}.
Language models heavily rely on semantic priors from pretraining~\citep{wei2023larger}, which hurts their ability to perform inductive reasoning—a core feature of human intelligence~\citep{peirce1868questions}.
Inductive reasoning involves identifying general principles from specific examples and applying them to novel situations. In an ICL scenario, the language model must first generate a hypothesis that captures the input-output relationship in the demonstrations and then apply this hypothesis to a novel input to produce the desired output~\citep{wang2023hypothesis,qiu2023phenomenal}.
To achieve high accuracy, even very large models like GPT-4~\citep{openai2024gpt4technicalreport} often rely on advanced methods such as chain-of-thought (CoT)~\citep{li2025mirageevaluatingexplaininginductive}, hypothesis search~\citep{wang2023hypothesis}, and iterative hypothesis refinement~\citep{qiu2023phenomenal}.
Although the recent reasoning model DeepSeek-R1~\citep{guo2025deepseek} shows impressive reasoning ability via test-time scaling~\citep{openai2024openaio1card}, few-shot prompting consistently degrades its performance, indicating that ICL remains a challenge.

Although instruction-tuning can be used to improve the ICL capabilities of a language model, previous works in this area mostly rely on imitation learning~\citep{huang2022incontextlearningdistillationtransferring,alpaca,vicuna2023}. Imitation learning teaches a student model to directly mimic the outputs of a teacher model but fails to teach the underlying rules connecting inputs and outputs in complex reasoning tasks~\citep{mukherjee2023orcaprogressivelearningcomplex}. 
In this work, we aim to improve the ICL capabilities of language models on inductive reasoning tasks via a novel reasoning distillation technique. We propose \proposed, a reasoning distillation method specifically designed to enhance the inductive reasoning abilities of language models. 
The main challenge in reasoning distillation is that the underlying rules are often not available as ground truth. We address this issue by using a teacher model to generate a set of candidate hypotheses to approximate the underlying rules. We also demonstrate that noisy fitness estimation—using natural language rule-following with the teacher model—can be effectively leveraged to evaluate the quality of rules, thereby improving model tuning.
\proposed consists of three key steps: (1) a data augmentation step, where we generate datasets for both rule generation and rule-following components of inductive reasoning using a teacher model; (2) a supervised fine-tuning (SFT) step, where the student model learns to mimic the rule generation and rule-following behavior of the teacher; and (3) an alignment step, which prepares the student model for inference-time optimizations for advanced inductive reasoning methods such as hypothesis search.

We find that \proposed helps small models like LLaMA-3, Mistral, and Qwen outperform proprietary models on inductive reasoning tasks, and even surpass the teacher model, GPT-4o~\citep{openai2024gpt4ocard}, in some cases.
We investigate the key steps and factors affecting the distillation process. Specifically, we show that the quality of a generated hypothesis can be evaluated by checking the number of demonstrated ICL training examples that satisfy that hypothesis. This measurement allows us to create a preference alignment corpus for rule generation and to filter the SFT corpus, both of which positively impact performance.
Finally, we examine the effects of different alignment techniques and observe that Odds Ratio Preference Optimization (ORPO)~\citep{hong2024orpomonolithicpreferenceoptimization} achieves the best overall performance compared to Direct Preference Optimization (DPO)~\citep{rafailov2024directpreferenceoptimizationlanguage} and Kahneman-Tversky Optimization (KTO)~\citep{ethayarajh2024ktomodelalignmentprospect} in our setting. We find that the preference alignment stage can significantly improve the inference-time efficiency of inductive reasoning.

Our contributions are as follows.
\begin{itemize}
    \item We propose \proposed, a reasoning distillation method specifically designed to improve the inductive reasoning ability of language models. \proposed demonstrates enhanced performance on four diverse tasks: 1D-ARC~\citep{xu2024llmsabstractionreasoningcorpus}, List Function~\citep{rule2020child}, ACRE~\citep{zhang2021acreabstractcausalreasoning}, and MiniSCAN~\citep{lake2019humanfewshotlearningcompositional}. Notably, the proposed approach outperforms even the teacher model, GPT-4o, on three out of four tasks within an equivalent hypothesis search space. \proposed, based on the LLaMA-3 backbone, achieves relative improvements of 23.2\%, 2.8\%, and 66.6\% over GPT-4o on 1D-ARC, ACRE, and MiniSCAN, respectively.
    
    \item We demonstrate the impact of different combinations of model tuning and alignment techniques on overall performance. We also perform inference-time optimization by aligning the models to generate higher-quality hypotheses within a smaller candidate search space compared to the teacher. \proposed, based on the LLaMA-3 backbone, performs equal to or better than the teacher model while being 87\%, 53\%, and 25\% more token-efficient on MiniSCAN, 1D-ARC, and ACRE, respectively.
\end{itemize}

\section{Methodology}
\label{sec:method}
In-context learning with few-shot prompting involves prompting a model with an instruction $\phi$ and $n$ demonstrations of input-output pairs $\mathcal{E} = \{(x_i, y_i)\}_{i=1}^n$, where $x_i \in X,\ y_i \in Y$. Here, $(x_i, y_i)$ denotes the input-output pair for the $i$-th demonstration, and $X$ and $Y$ refer to the sets of inputs and outputs, respectively. The model is expected to exploit the demonstrated examples to produce an output $y'$ for an unseen input $x'$. In the inductive reasoning variant of ICL, the model is prompted with a rule-generation instruction $\phi_{\mathrm{rg}}$ to learn a function $f: X \rightarrow Y$ that maps inputs to outputs. We refer to this step as the \emph{rule or hypothesis generation} step. Next, the model is prompted with an instruction $\phi_{\mathrm{rf}}$, the learned rule $f$, and an unseen input $x'$ to produce the output $y'$. We refer to this step as \emph{rule-following}. Our objective is to distill the inductive reasoning ability of a teacher model into a student model.

\begin{figure}
    \centering
    \includegraphics[width=\textwidth]{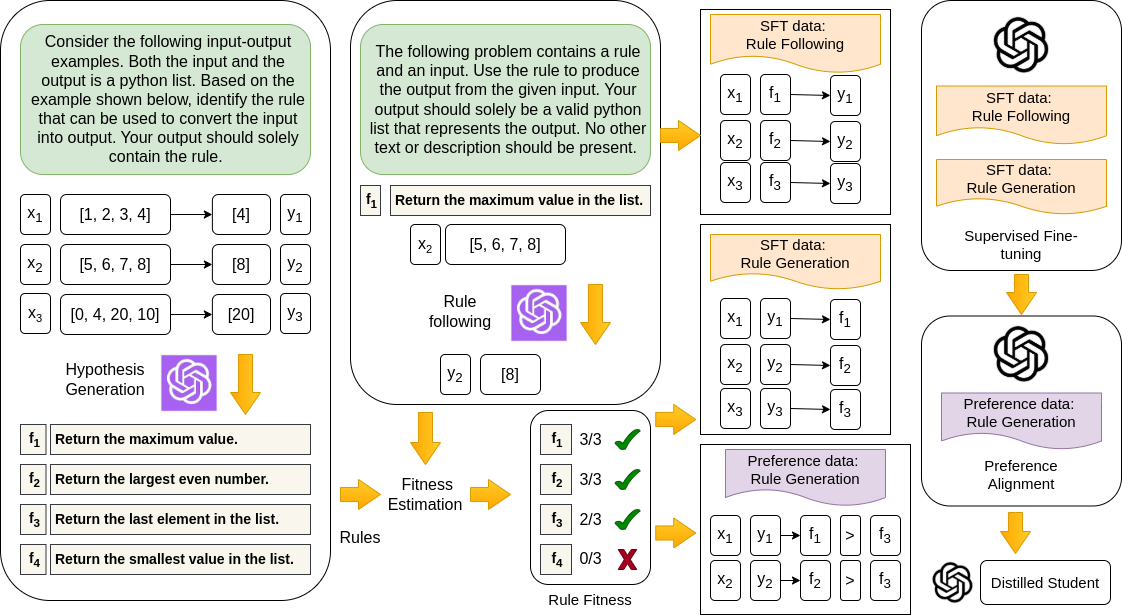}
    \caption{Overview of data augmentation, filtering and model tuning in \proposed. The teacher and student models are shown in purple and black, respectively. Each entry in the rule generation corpus contains multiple few-shot demonstrations. Each entry in the SFT and preference alignment corpora includes both the rule generation and the corresponding rule-following instructions. These details are omitted for clarity.}
    \label{fig:proposed_method}
    \vspace{-1em}
\end{figure}

Some reasoning tasks, such as the 1D ARC task, are difficult for open-weight models, and prior works have therefore used large proprietary models such as GPT-4~\citep{wang2023hypothesis}. This prompted us to use GPT-4o as our teacher model. Using a proprietary model as a teacher poses challenges for using traditional distillation techniques that rely on output logits~\citep{hinton2015distillingknowledgeneuralnetwork} or intermediate representations~\citep{romero2015fitnetshintsdeepnets}. To mitigate this issue, we use a black-box distillation approach via data augmentation~\citep{li2022explanationslargelanguagemodels}. Our method has three steps: data augmentation using the teacher model, supervised fine-tuning, and preference alignment of the student model. An overview of this process is shown in Figure~\ref{fig:proposed_method}.

\paragraph{Data augmentation} 
During data augmentation, we first use a set of in-context demonstrations, $\mathcal{E}$, to generate a set of $m$ hypotheses from the teacher model. Let $f_j$ be one candidate hypothesis. The next step is to use the teacher model to perform rule following and determine how well the examples in $\mathcal{E}$ align with $f_j$. Thus, a fitness score $s_j$ is computed for a candidate hypothesis $f_j$ as follows:
\begin{equation*}
    s_j = \sum_{i=1}^n [f_j(x_i) = y_i]
\end{equation*}
Since the teacher model's rule-following ability is imperfect, we refer to the process as noisy fitness estimation. After data augmentation, we perform quality analysis and filtering. We construct two separate datasets: one for supervised fine-tuning involving both rule following and rule generation, and another for preference alignment containing chosen and rejected rule pairs.

Let $T$ be the number of reasoning tasks in the ICL dataset. Let $\Gamma_{\mathrm{rg}}^{\mathrm{s}}$, $\Gamma_{\mathrm{rf}}^{\mathrm{s}}$, and $\Gamma_{\mathrm{rg}}^{\mathrm{p}}$ refer to the SFT dataset for rule generation, the SFT dataset for rule following, and the preference alignment dataset for rule generation, respectively. They are constructed as follows:
\begin{align*}
    \Gamma_{\mathrm{rg}}^{\mathrm{s}} &= \left\{ (\phi_{\mathrm{rg}}, \mathcal{E}^t, f_j^t) : s_j^t \geq \tfrac{n}{2} \right\}_{t=1}^T \\
    \Gamma_{\mathrm{rf}}^{\mathrm{s}} &= \left\{ (\phi_{\mathrm{rf}}, f_j^t, x_i^t, y_i^t) : s_j^t \geq \tfrac{n}{2},\; 1 \leq i \leq n \right\}_{t=1}^T \\
    \Gamma_{\mathrm{rg}}^{\mathrm{p}} &= \left\{ (\phi_{\mathrm{rg}}, \mathcal{E}^t, f_j^t, f_k^t) : 1 \leq j,k \leq m,\; s_j^t > s_k^t + d \right\}_{t=1}^T
\end{align*}
To ensure the high quality of the SFT datasets, we only allow hypotheses that satisfy at least half of the examples in $\mathcal{E}$. To check whether a hypothesis satisfies an example, we use the teacher model's rule-following ability with natural language. Since the teacher model may not perfectly follow rules, even a perfect hypothesis might not receive a perfect score. Setting the rule quality threshold to $50\%$ ensures that the high-quality rules remain in the dataset. For the preference dataset, we select all hypothesis pairs within a task such that the score difference between the chosen rule and the rejected rule is greater than a threshold $d$.

\paragraph{Supervised Fine-tuning} We perform supervised fine-tuning on the student model during the model tuning phase using low-rank adaptation. The SFT dataset comprises both rule generation and rule-following examples across all tasks. The goal is to maximize the likelihood $\log P(f_j \mid \phi_{\mathrm{rg}}, \mathcal{E})$ and $\log P(y_i \mid \phi_{\mathrm{rf}}, f_j, x_i)$ for rule generation and rule following, respectively.

\paragraph{Preference Alignment} 
After supervised fine-tuning, we perform preference alignment to improve the quality of rule generation within a smaller hypothesis search space. The best-performing alignment method, ORPO, is defined based on the odds ratio of generating the chosen hypothesis relative to the rejected hypothesis. Given an instruction $\phi_{\mathrm{rg}}$ and ICL demonstration $\mathcal{E}$, the odds of generating a hypothesis $f_j$ are defined as:
\begin{align*}
    \mathrm{odds}(f_j \mid \phi_{\mathrm{rg}}, \mathcal{E}) 
    &= \frac{P(f_j \mid \phi_{\mathrm{rg}}, \mathcal{E})}{1 - P(f_j \mid \phi_{\mathrm{rg}}, \mathcal{E})}
\end{align*}

The odds ratio of generating the chosen hypothesis $f_j$ relative to the rejected hypothesis $f_k$, and the associated loss function $\mathcal{L}_{\mathrm{or}}$, are as follows:
\begin{align*}
    \mathrm{odds\_ratio}(f_j, f_k) 
    &= \frac{\mathrm{odds}(f_j \mid \phi_{\mathrm{rg}}, \mathcal{E})}{\mathrm{odds}(f_k \mid \phi_{\mathrm{rg}}, \mathcal{E})} \\
    \mathcal{L}_{\mathrm{or}} 
    &= -\log \sigma\big( \log \mathrm{odds\_ratio}(f_j, f_k) \big)
\end{align*}

The overall preference alignment loss for ORPO is computed via a weighted combination of the SFT loss $\mathcal{L}_{\mathrm{sft}}$ and the odds ratio loss $\mathcal{L}_{\mathrm{or}}$:
\begin{align*}
    \mathcal{L} 
    &= \mathcal{L}_{\mathrm{sft}} + \lambda \mathcal{L}_{\mathrm{or}}
\end{align*}

\section{Experiments}
\label{sec:experiment}
\subsection{Datasets}
We evaluate our system on several datasets. These datasets offer a wide range of inductive reasoning tasks such as 1D grid manipulation, list operations, causal induction abilities, and sequence-to-sequence operations. We randomly select 10\% of the tasks from each dataset for testing and use the rest for constructing an augmented dataset for model tuning. The data augmentation is performed with GPT-4o as the teacher model, using a hypothesis search size of 50. The statistics of the training data are shown in Table~\ref{tab:dataset_summary}. Details regarding the hypothesis quality within the dataset is shown in Figure~\ref{fig:data_quality} in Appendix~\ref{app:data_quality}.

\begin{table}
\small
\centering
\begin{tabular}{@{}lrrrr@{}}
\toprule
\textbf{} & \textbf{Preference Data} & \textbf{SFT Data} & \textbf{SFT Data} & \textbf{SFT Data} \\ \midrule
\textbf{Tasks} & \textbf{Rule Generation} & \textbf{Rule Generation} & \textbf{Rule Following} & \textbf{Direct Few-shot} \\ \midrule
List Function & 31,186 & 10,514 & 22,568 & 1,800 \\
1D ARC & 38,717 & 20,250 & 17,661 & 810 \\
ACRE & 19,461 & 4,472 & 17,760 & 360 \\ 
SCAN & 19,404 & 4,500 & 30,380 & 900 \\ \midrule
Total & 108,768 & 39,736 & 88,369 & 3,870 \\ \bottomrule
\end{tabular}
\caption{Training Data Statistics}
\label{tab:dataset_summary}
\vspace{-1em}
\end{table}

\paragraph{List Function} \citet{rule2020child} proposed the List Function dataset to provide insight into children's learning abilities from a computer programming perspective. The dataset contains 250 tasks, with both input and output examples provided. Each task comprises eight training and eight test examples, making them suitable for few-shot prompting scenarios. The list operations that map inputs to outputs include operations such as selection, duplication, sorting, and more.

\paragraph{1D ARC} The Abstraction and Reasoning Corpus (ARC)~\citep{chollet2019measureintelligence} is a dataset designed to evaluate machine learning models on complex reasoning and generalization tasks. Each problem in ARC consists of a set of input-output grid-based examples ($30\times30$) and a test input grid. The goal is to infer the transformation pattern and apply it to the test input in order to generate the correct output. The 1D ARC dataset~\citep{xu2024llmsabstractionreasoningcorpus} is a simplified variant of ARC where problems are structured in a one-dimensional format instead of a 2D grid. It retains the core principles of ARC but restricts the input and output examples to linear sequences ($1\times30$) rather than spatial grids.

\paragraph{ACRE} The Abstract Causal REasoning (ACRE)~\citep{zhang2021acreabstractcausalreasoning} dataset is a benchmark designed to evaluate and enhance the causal induction capabilities of visual reasoning systems. Inspired by the ``Blicket'' experiments from developmental psychology, ACRE assesses how well AI models can infer causal relationships beyond mere statistical associations.

\paragraph{MiniSCAN} MiniSCAN~\citep{lake2019humanfewshotlearningcompositional} is a sequence-to-sequence task where an input sequence corresponding to the instruction is transformed into an output action sequence. The instruction is composed of primitive pseudo-words, and the output action sequence consists of colored circles.

\subsection{Baselines \& Metrics}

\paragraph{Baseline Input-to-Output (IO)} This baseline involves a simple few-shot prompting approach without any intermediate reasoning steps. Prompting is performed on frozen language models. Decoding is performed with a temperature of $0.7$ and a top-p of $1.0$. The prompts associated with direct few-shot prompting are shown in Table~\ref{tab:prompts} in Appendix~\ref{app:prompts}. 

\paragraph{Baseline Hypothesis Search (HS)} Following \citet{wang2023hypothesis}, we use the hypothesis search on frozen language models as a baseline. The first step is to generate multiple hypotheses for inductive reasoning that may explain the demonstrated ICL examples. The hypothesis that best fits the examples is chosen. The second step involves applying the hypothesis to a novel input to produce the output. The decoding temperature for rule generation and rule following is $0.9$ and $0.7$, respectively, with top-p set to $1.0$. The prompts associated with hypothesis search are shown in Table~\ref{tab:prompts} in Appendix~\ref{app:prompts}. 

\paragraph{Baseline Input-to-Output with SFT} This baseline involves simple few-shot prompting using models that have been fine-tuned in a multi-task setting. We fine-tune the base models using the data presented in the last column of Table~\ref{tab:dataset_summary}. Training is performed using LoRA (rank $8$, scaling factor $16$). We tune the models for 3 epochs with a learning rate of $10^{-4}$ using the Adam optimizer. The decoding temperature during inference is $0.7$ with a top-p of $1.0$. 

\paragraph{Metrics} 
Raw accuracy is used as the evaluation metric for performance. It is defined as the fraction of correctly predicted ICL outputs with respect to the total number of outputs in the test set. We use the cost-to-performance ratio (CP) as a metric to measure efficiency. CP is defined as the ratio of average token consumption per output to accuracy.

\subsection{Training Details}
We use three language models for our experiments: Llama-3-8B-Instruct~\citep{grattafiori2024llama3herdmodels}, Mistral-7B-Instruct-v0.3~\citep{jiang2023mistral7b}, and Qwen2.5-7B-Instruct~\citep{bai2023qwentechnicalreport}. We perform supervised fine-tuning on both rule generation and rule following, and then perform alignment on preferred rule generation. Supervised fine-tuning is performed using LoRA adaptation with a rank of $8$ and a scaling factor of $16$. The models are fine-tuned for 4 epochs with a learning rate of $10^{-4}$. Alignment is also performed using LoRA adaptation with a rank of $8$ and a scaling factor of $16$. During alignment, we fine-tune the models for 5 epochs with a learning rate of $5\times 10^{-6}$. Training is performed using $2\times 48$~GB Nvidia RTX A6000 GPUs with a per-device batch size of $1$ and gradient accumulation steps of $8$. The durations of SFT and alignment are 48 and 72 hours, respectively. As shown in Table~\ref{tab:main_result}a, we experiment with different combinations of SFT and alignment for Llama-3-8B-Instruct. We then apply the best-performing combination to the two remaining models.

During inference with \proposed, we set a decoding temperature of $0.9$ for rule generation and $0.7$ for rule following, with a top-p set to $1.0$. The best decoding temperatures are estimated with a grid search, as shown shown in Figure~\ref{fig:temperature_search} in Appendix~\ref{app:temperature}. Unless otherwise specified, we use the same model checkpoint for rule generation and rule following during inference.

\begin{table}
\resizebox{\textwidth}{!}{
\begin{tabular}{lllrrrrr}
\toprule
& \textbf{Method} & \textbf{Tuning} & \textbf{List Func} & \textbf{1D ARC} & \textbf{ACRE} & \textbf{MiniSCAN} \\ \midrule
\multirow{5}{*}{a)} & Baseline IO & N/A & $0.43\pm0.01$ & $0.07\pm 0.01$ & $0.53\pm 0.01$ & $0.10\pm 0.01$ \\
& Baseline IO & SFT (Few-shot) &  $0.47\pm 0.01$ &  $0.40\pm 0.01$ &  $0.70\pm 0.01$ &  $0.25\pm 0.01$ \\
& Baseline HS & N/A &  $0.36\pm 0.01$ &  $0.04\pm 0.01$ &  $0.08\pm 0.01$ &  $0.00\pm 0.00$ \\
& \multirow{5}{*}{\proposed-Llama} & SFT (RG+RF)\footnotemark[1] &  $0.43\pm 0.00$ &  $\underline{0.47}\pm 0.00$ &  $0.65\pm 0.00$ &  $0.21\pm 0.00$ \\
&  & SFT (RG+RF) + DPO (RG) &  $\textbf{0.49}\pm 0.01$ &  $\underline{0.51}\pm 0.01$ &  $\underline{\textbf{0.73}}\pm 0.01$ &  $\underline{0.42}\pm 0.01$ \\
& & SFT (RG+RF) + ORPO (RG)\footnotemark[1] & $\textbf{0.49}\pm 0.01$ & $\underline{\textbf{0.53}}\pm 0.01$ & $\underline{\textbf{0.73}}\pm 0.01$ & $\underline{\textbf{0.45}}\pm 0.01$ \\
&  & SFT (RG+RF)+ KTO (RG) &  $0.48\pm 0.01$ &  $\underline{0.51}\pm 0.01$ &  $0.72\pm 0.01$ &  $\underline{0.38}\pm 0.01$ \\
& & SFT (RF) + ORPO (RG) &  $0.41\pm 0.01$ &  $0.43\pm 0.01$ &  $0.59\pm 0.01$ &  $0.21\pm 0.01$ \\ \midrule 
\multirow{5}{*}{b)} & Baseline IO & N/A & $0.43\pm0.00$ & $0.16\pm 0.01$ & $0.70\pm 0.00$ & $0.04\pm 0.00$ \\
& Baseline IO & SFT (Few-shot) &  $0.44\pm 0.00$ &  $0.17\pm 0.00$ &  $0.73\pm 0.00$ &  $0.25\pm 0.00$ \\
& Baseline HS & N/A & $0.33\pm 0.01$ &  $0.01\pm 0.01$ &  $0.00\pm 0.00$ &  $0.02\pm 0.01$ \\
&\multirow{2}{*}{\proposed-Qwen} & SFT (RG+RF)\footnotemark[1] &  $0.45\pm 0.00$ &  $\underline{0.45}\pm 0.00$ &  $0.68\pm 0.00$ &  $0.19\pm 0.00$ \\
& & SFT (RG+RF) + ORPO (RG)\footnotemark[1] & $\textbf{0.54}\pm 0.01$ & $\underline{\textbf{0.48}}\pm 0.01$ & $\underline{\textbf{0.80}}\pm 0.01$ & $\underline{\textbf{0.43}}\pm 0.01$ \\ \midrule
\multirow{5}{*}{c)} & Baseline IO & N/A & $0.41\pm0.00$ & $0.09\pm 0.01$ & $0.52\pm 0.00$ & $0.08\pm 0.00$ \\
& Baseline IO & SFT (Few-shot) &  $0.42\pm 0.00$ &  $0.27\pm 0.00$ &  $0.75\pm 0.00$ &  $0.22\pm 0.00$ \\
& Baseline HS & N/A &  $0.25\pm 0.01$ &  $0.02\pm 0.01$ &  $0.05\pm 0.01$ &  $0.00\pm 0.00$ \\
&\multirow{2}{*}{\proposed-Mistral} & SFT (RG+RF)\footnotemark[1] &  $0.41\pm 0.00$ &  $0.40\pm 0.00$ &  $0.70\pm 0.00$ &  $0.10\pm 0.00$ \\
& & SFT (RG+RF) + ORPO (RG)\footnotemark[1] & $\textbf{0.45}\pm 0.01$ & $\underline{\textbf{0.53}}\pm 0.01$ & $\underline{\textbf{0.78}}\pm 0.01$ & $\underline{\textbf{0.35}}\pm 0.01$ \\ \midrule
\multicolumn{2}{c}{gpt-3.5-turbo\footnotemark[2] IO} & N/A &  $0.49\pm 0.01$ &  $0.14\pm 0.01$ &  $0.00\pm 0.00$ &  $0.01\pm 0.01$ \\
\multicolumn{2}{c}{gpt-3.5-turbo\footnotemark[2] HS} & N/A &  $0.50\pm 0.01$ &  $0.08\pm 0.01$ &  $0.00\pm 0.00$ &  $0.01\pm 0.01$ \\ \midrule
\multicolumn{2}{c}{deepseek v3 IO} & N/A &  $0.68\pm 0.01$ &  $0.67\pm 0.01$ &  $0.72\pm 0.00$ &  $0.26\pm 0.01$ \\ 
\multicolumn{2}{c}{deepseek v3 HS} & N/A &  $0.61\pm 0.01$ &  $0.40\pm 0.01$ &  $0.00\pm 0.00$ &  $0.04\pm 0.01$ \\ \midrule
\multicolumn{2}{c}{gpt-4o IO} & N/A &  $0.61\pm 0.01$ &  $0.44\pm 0.01$ &  $0.77\pm 0.01$ &  $0.27\pm 0.01$ \\ 
\multicolumn{2}{c}{gpt-4o HS} & N/A &  $0.69\pm 0.01$ &  $0.43\pm 0.01$ &  $0.71\pm 0.01$ &  $0.27\pm 0.01$ \\ \bottomrule
\end{tabular}
}
\caption{Performance of \proposed is evaluated using the following models: (a) Llama-3-8B-Instruct, (b) Qwen2.5-7B-Instruct, and (c) Mistral-7B-Instruct. IO refers to direct input-to-output few-shot prompting, and HS refers to the hypothesis search method. Both the hypothesis search baselines and \proposed use a hypothesis size of 10. RG and RF refer to rule generation and rule following, respectively. Underlined results indicate cases where a variant of \proposed outperforms GPT-4o with hypothesis search.
}
\label{tab:main_result}
\end{table}

\footnotetext[1]{We will refer to the SFT (RG+RF) + ORPO (RG) variant of \proposed model simply as \proposed and SFT (RG+RF) variant of \proposed model as SFT.}
\footnotetext[2]{gpt-3.5-turbo-1106}

\subsection{Results}
\paragraph{Comparison with baselines} Table~\ref{tab:main_result} shows the performance of \proposed compared to three baselines, across several open-weight models. \proposed outperforms the three baselines for each base model. In particular, \proposed performs better than the strongest few-shot prompting baseline, which has been fine-tuned for the few-shot ICL setting on the same tasks. For example, \proposed-Qwen achieves 23\%, 182\%, 9\%, and 72\% relative improvement over Baseline IO-SFT on List Function, 1D ARC, ACRE, and MiniSCAN, respectively. This demonstrates that reasoning distillation, which teaches the student model the underlying rules, is more effective than direct input-output ICL distillation, as the latter may not generalize well to novel inputs.

\paragraph{Comparison with the teacher} The \proposed model outperforms the teacher model, GPT-4o, on three out of four tasks across all three base models. For example, \proposed-Llama achieves relative improvements of 23.2\%, 2.8\%, and 66.6\% over GPT-4o for 1D ARC, ACRE, and MiniSCAN, respectively, when both models use a hypothesis search size of 10. This improvement can be attributed to the difference in hypothesis size during data augmentation and inference, as well as the impact of data filtering, SFT, and alignment. We performed data augmentation with a hypothesis size of 50. During data filtering, some low-quality rules in the candidate space are filtered out. Thus, the SFT stage helps the student model generate higher-quality rules compared to the teacher within an equivalent candidate set. The alignment stage further ensures that the best candidates are among the top candidates generated by the distilled model. Consequently, the distilled model with a smaller hypothesis search space can mimic the performance of the teacher model with a larger hypothesis search space.

\paragraph{Impact of SFT and Alignment Combinations} For \proposed-Llama, we first perform SFT on rule generation and rule-following. We then apply three preference alignment strategies to the fine-tuned model individually: DPO, ORPO, and KTO. We find that ORPO achieves the best performance, outperforming DPO on 1D ARC and MiniSCAN by two and three percentage points, respectively. While the SFT loss during initial fine-tuning jointly optimizes both rule generation and rule-following, the SFT loss within ORPO exclusively optimizes rule generation. Therefore, we hypothesize that the SFT loss component in ORPO contributes to the additional improvement. Since \citet{hong2024orpomonolithicpreferenceoptimization} originally proposed ORPO to circumvent the need for a separate SFT stage, we experiment with a variant in which SFT is performed on rule-following only. ORPO is then applied to rule generation (the last row of Table~\ref{tab:main_result}a). We find that this variant underperforms even compared to the SFT-only version. This indicates that if the discrepancy between the rule quality generated by the teacher and the base student model is too large, ORPO alone cannot eliminate the need for fine-tuning on rule generation.

\begin{figure}
\centering
\includegraphics[width=\textwidth]{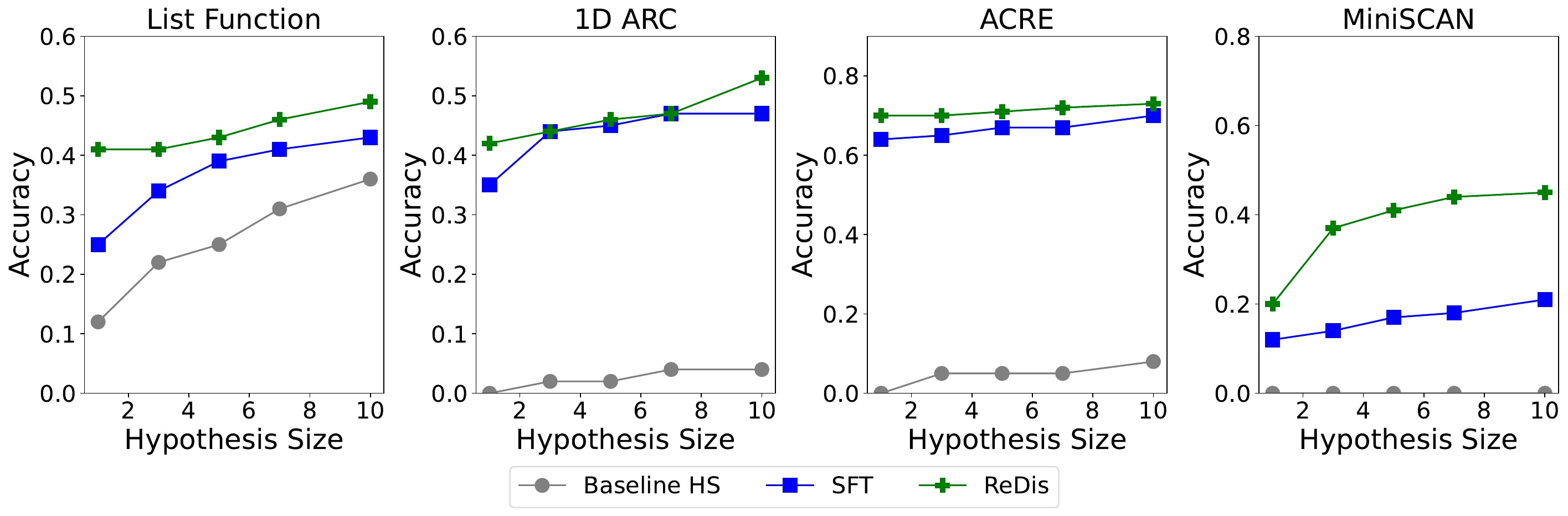}
\caption{The impact of hypothesis size on performance improvement of \proposed-Llama. The results are shown for hypothesis size of $1, 3, 5, 7,$ and $10$.}
\label{fig:accuracy_hypo}
\end{figure}

\paragraph{Impact of Hypothesis Size} Figure~\ref{fig:accuracy_hypo} shows the impact of hypothesis size on \proposed-Llama performance. We find that the aligned model performs better than the base model and the SFT-only variant across all hypothesis sizes. The performance gap is more prominent for a very small candidate set, as demonstrated in the case of List Function and 1D ARC tasks for the hypothesis size of one. This indicates that preference alignment can help the student model generate better rules at the top of the candidate list.

\begin{table}
\small
\centering
\begin{tabular}{lrrrrrrrrr}
\toprule
Models & $m$ & \multicolumn{2}{c}{List Func} & \multicolumn{2}{c}{1D ARC} & \multicolumn{2}{c}{ACRE} & \multicolumn{2}{c}{MiniSCAN} \\ \midrule
&  & Acc\textuparrow & CP\textdownarrow & Acc\textuparrow & CP\textdownarrow & Acc\textuparrow & CP\textdownarrow & Acc\textuparrow & CP\textdownarrow \\
 \midrule
Baseline & 10 & 0.36 & 72 & 0.04 & 73050 & 0.08 & 8287 & 0.00 & -- \\
gpt-3.5-turbo &10 & 0.50 & 469 & 0.08 & 19321 & 0.00 & -- & 0.01 & 27200 \\
gpt-4o &10 & 0.69 & 646 & 0.43 & 8317 & 0.71 & 235 & 0.27 & 4925 \\
deepseek v3 &10 & 0.61 & 1013 & 0.40 & 13272 & 0.00 & -- & 0.04 & 55525 \\
SFT &10 & 0.43 & 625 & 0.47 & 7493 & 0.65 & 210 & 0.21 & 2704 \\
\proposed &10 & 0.49 & \underline{644} & 0.53 & \underline{6611} & 0.73 & 338 & 0.45 & \underline{1095} \\ 
\proposed &5 & 0.43 & \underline{\textbf{393}} & 0.46 & \underline{\textbf{3869}} & 0.71 & \underline{\textbf{176}} & 0.41 & \underline{\textbf{631}} \\ 
\bottomrule
\end{tabular}
\caption{Cost to performance (CP) ratio of \proposed-Llama is compared with the base, fine-tuned, and proprietary models. All models use hypothesis search and $m$ denotes the size of the search space. \proposed variants that outperform GPT-4o is underlined.}
\label{tab:token_consumption}
\end{table}

\paragraph{Inference-time optimization} One significant advantage of the proposed reasoning distillation is the inference-time optimizations it can provide over the teacher model. Table~\ref{tab:token_consumption} shows the cost-to-performance ratio for each test set. \proposed-Llama with top-5 hypotheses achieves 87\%, 53\%, and 25\% relative improvements in cost-to-performance ratio compared to the teacher model for MiniSCAN, 1D ARC, and ACRE, respectively.

\paragraph{Impact of hypothesis length} Interestingly, we find that the \proposed model trained only with SFT sometimes becomes more token-efficient than the teacher model, even when the hypothesis size is the same. For example, the SFT-only variant is 10\% more token-efficient than GPT-4o on the 1D ARC task (shown in Table~\ref{tab:token_consumption}). To investigate this observation, we calculate the Pearson correlation coefficient between hypothesis length and quality. The task- and dataset-specific correlation coefficients are reported in Table~\ref{tab:hypo_len}. We find that longer hypotheses tend to have lower quality scores for the List Function and 1D ARC tasks. It is possible that language models may struggle to follow overly long or complex rules. Therefore, longer hypotheses are automatically removed from the SFT dataset during fitness estimation, and the fine-tuned model prefers shorter hypotheses that are easier to follow.

\begin{table}
\small
\centering
\begin{tabular}{lrrrrr}
\toprule
Attribute & All & List Func & 1D ARC & ACRE & MiniSCAN \\ \midrule
Pearson correlation coefficient & -0.20 & -0.31 & -0.25 & 0.17 & 0.04 \\
Share of SFT Data & 1.00 & 0.26 & 0.51 & 0.11 & 0.12 \\ \bottomrule
\end{tabular}
\caption{Pearson correlation coefficients between hypothesis quality and hypothesis length are shown for all SFT data and task-specific SFT data. We find that longer hypotheses may be associated with lower overall quality in most cases.}
\label{tab:hypo_len}
\end{table}

\begin{table}
\resizebox{\textwidth}{!}{
\begin{tabular}{llcccc}
\toprule
RG Model & RF Model & List Func & 1D ARC & ACRE & MiniSCAN \\ \midrule
SFT (RG+RF) + ORPO (RG) & SFT (RG+RF) + ORPO (RG) & 0.49 & 0.53 & 0.73 & 0.45 \\
SFT (RG+RF) + ORPO (RG) & Base & 0.43 & 0.12 & 0.08 & 0.01 \\
SFT (RG+RF) + ORPO (RG) & SFT (RG+RF) & 0.49 & 0.54 & 0.73 & 0.46 \\
Base & SFT (RG+RF) + ORPO (RG) & 0.39 & 0.17 & 0.55 & 0.09 \\
SFT (RG+RF) & SFT (RG+RF) + ORPO (RG) & 0.45 & 0.46 & 0.68 & 0.21 \\ \bottomrule
\end{tabular}
}
\caption{Ablation Study for \proposed-Llama, for hypothesis size of $10$.}
\label{tab:ablation}
\end{table}

\paragraph{Ablation Study} During the ablation study, we use separate checkpoints for rule generation and rule following to determine the impact of SFT and alignment on each reasoning step separately, as shown in Table~\ref{tab:ablation}. The first row indicates our default setup, where the same model is used for both reasoning steps. The most significant performance drop occurs when rule following is performed with a base LLaMA model. For example, we see a 77\% relative drop in accuracy for the 1D ARC task. This indicates that fine-tuning on rule following is critical for optimal performance, as this allows the system to evaluate and select the best hypothesis during hypothesis search. We also find that performing alignment on rule generation with a low-rank adaptation does not degrade the rule-following ability of the models, as the performance is comparable to the ablation scenario where the SFT-only model is used for rule following. We see a 68\% drop in accuracy when the base model is used for rule generation and a 13\% drop when only alignment is removed for the rule generation model. Therefore, SFT on rule generation is the second most critical factor in \proposed's performance, followed by alignment on rule generation.

\section{Related Work}
\label{sec:related}
\paragraph{Inductive Reasoning with Language Models} In the context of language models, inductive reasoning involves generating rules from observed examples that capture the underlying data-generating mechanism~\citep{wang2023hypothesis, qiu2023phenomenal, yang-etal-2024-language}, with diverse downstream applications in domains such as coding~\citep{shao-etal-2025-case2code} and dialogue systems~\citep{xie-etal-2024-shot-dialogue}. While several recent works benchmark and evaluate language models' inductive reasoning abilities~\citep{li2025mirageevaluatingexplaininginductive, gendron2024largelanguagemodelsstrong}, how to systematically enhance the inductive reasoning capabilities of language models, as studied in this work, remains an open problem. While existing works aim to improve inductive reasoning using frozen models through hypothesis search~\citep{wang2023hypothesis} and refinement~\citep{qiu2023phenomenal}, these methods rely on large proprietary models, which can be prohibitively expensive for inference. We demonstrate that systematically improving the inductive reasoning ability of open-weight models through distillation can match the accuracy of large proprietary models while allowing significantly cheaper inference.

\paragraph{Distillations for Language Model Training} 
Recently, there has been increasing interest in distilling domain-specific or task-specific knowledge and abilities from large language models. This paradigm has been applied to domains such as summarization~\citep{jung-etal-2024-impossible}, commonsense reasoning~\citep{west-etal-2022-symbolic}, chain-of-thought reasoning~\citep{li-etal-2024-mode}, among others. However, it is unclear whether this paradigm can be applied to the distillation of inductive reasoning abilities from language models, especially if the quality of intermediate reasoning steps generated by the model is not evaluated. We address this issue in our work and demonstrate that effective evaluation of the hypothesis-generation step can lead to improved accuracy and reduced cost for the resulting distilled models.

\paragraph{Learning from In-context Demonstrations}
ICL has emerged as a powerful paradigm for language model reasoning~\citep{dong-etal-2024-survey}. A growing body of research highlights the significance of demonstration selection for ICL. Prior work finds that the number, order, and diversity of in-context examples significantly affect model performance~\citep{min-etal-2022-rethinking, levy-etal-2023-diverse, peng-etal-2024-revisiting}. Retrieval has been shown to be an effective method for demonstration selection~\citep{li-etal-2023-unified, DBLP:conf/nips/ZhangZ023, DBLP:journals/corr/abs-2401-11624, gao2025learning}. Additionally, many studies propose methods to enhance the adaptability and generalization of ICL~\citep{metaICL,DBLP:conf/icml/LiIPO23,xu-etal-2023-unleash, wei-etal-2023-symbol}.

\section{Conclusion}
We propose \proposed, a black-box reasoning distillation method to improve the inductive reasoning abilities of language models. We use the teacher model to augment the rule generation and rule-following dataset. Then, we perform data filtering by checking the fitness of the generated rule based on the demonstrated ICL examples. We perform supervised fine-tuning on the reasoning steps, followed by alignment to help the model generate high-quality hypotheses within a reduced candidate search space. \proposed outperforms equivalent few-shot prompting baselines and achieves relative improvements of 23.2\%, 2.8\%, and 66.6\% over the teacher model in 1D ARC, ACRE, and MiniSCAN. \proposed also reduces token consumption, with relative improvements of 87\%, 53\%, and 25\% in the cost-to-performance ratio over the teacher on MiniSCAN, 1D-ARC, and ACRE. An interesting direction for future research is to address reasoning tasks with ambiguous rules where the few-shot demonstrations are insufficient to pinpoint a single underlying rule. %

\bibliography{colm2025_conference}

\begin{thebibliography}{49}
\providecommand{\natexlab}[1]{#1}
\providecommand{\url}[1]{\texttt{#1}}
\expandafter\ifx\csname urlstyle\endcsname\relax
  \providecommand{\doi}[1]{doi: #1}\else
  \providecommand{\doi}{doi: \begingroup \urlstyle{rm}\Url}\fi

\bibitem[Bai et~al.(2023)Bai, Bai, Chu, Cui, Dang, Deng, Fan, Ge, Han, Huang, Hui, Ji, Li, Lin, Lin, Liu, Liu, Lu, Lu, Ma, Men, Ren, Ren, Tan, Tan, Tu, Wang, Wang, Wang, Wu, Xu, Xu, Yang, Yang, Yang, Yang, Yao, Yu, Yuan, Yuan, Zhang, Zhang, Zhang, Zhang, Zhou, Zhou, Zhou, and Zhu]{bai2023qwentechnicalreport}
Jinze Bai, Shuai Bai, Yunfei Chu, Zeyu Cui, Kai Dang, Xiaodong Deng, Yang Fan, Wenbin Ge, Yu~Han, Fei Huang, Binyuan Hui, Luo Ji, Mei Li, Junyang Lin, Runji Lin, Dayiheng Liu, Gao Liu, Chengqiang Lu, Keming Lu, Jianxin Ma, Rui Men, Xingzhang Ren, Xuancheng Ren, Chuanqi Tan, Sinan Tan, Jianhong Tu, Peng Wang, Shijie Wang, Wei Wang, Shengguang Wu, Benfeng Xu, Jin Xu, An~Yang, Hao Yang, Jian Yang, Shusheng Yang, Yang Yao, Bowen Yu, Hongyi Yuan, Zheng Yuan, Jianwei Zhang, Xingxuan Zhang, Yichang Zhang, Zhenru Zhang, Chang Zhou, Jingren Zhou, Xiaohuan Zhou, and Tianhang Zhu.
\newblock Qwen technical report.
\newblock \emph{CoRR}, abs/2309.16609, 2023.
\newblock \doi{10.48550/ARXIV.2309.16609}.
\newblock URL \url{https://doi.org/10.48550/arXiv.2309.16609}.

\bibitem[Brown et~al.(2020)Brown, Mann, Ryder, Subbiah, Kaplan, Dhariwal, Neelakantan, Shyam, Sastry, Askell, Agarwal, Herbert{-}Voss, Krueger, Henighan, Child, Ramesh, Ziegler, Wu, Winter, Hesse, Chen, Sigler, Litwin, Gray, Chess, Clark, Berner, McCandlish, Radford, Sutskever, and Amodei]{brown2020language}
Tom~B. Brown, Benjamin Mann, Nick Ryder, Melanie Subbiah, Jared Kaplan, Prafulla Dhariwal, Arvind Neelakantan, Pranav Shyam, Girish Sastry, Amanda Askell, Sandhini Agarwal, Ariel Herbert{-}Voss, Gretchen Krueger, Tom Henighan, Rewon Child, Aditya Ramesh, Daniel~M. Ziegler, Jeffrey Wu, Clemens Winter, Christopher Hesse, Mark Chen, Eric Sigler, Mateusz Litwin, Scott Gray, Benjamin Chess, Jack Clark, Christopher Berner, Sam McCandlish, Alec Radford, Ilya Sutskever, and Dario Amodei.
\newblock Language models are few-shot learners.
\newblock In Hugo Larochelle, Marc'Aurelio Ranzato, Raia Hadsell, Maria{-}Florina Balcan, and Hsuan{-}Tien Lin (eds.), \emph{Advances in Neural Information Processing Systems 33: Annual Conference on Neural Information Processing Systems 2020, NeurIPS 2020, December 6-12, 2020, virtual}, 2020.
\newblock URL \url{https://proceedings.neurips.cc/paper/2020/hash/1457c0d6bfcb4967418bfb8ac142f64a-Abstract.html}.

\bibitem[Chiang et~al.(2023)Chiang, Li, Lin, Sheng, Wu, Zhang, Zheng, Zhuang, Zhuang, Gonzalez, Stoica, and Xing]{vicuna2023}
Wei-Lin Chiang, Zhuohan Li, Zi~Lin, Ying Sheng, Zhanghao Wu, Hao Zhang, Lianmin Zheng, Siyuan Zhuang, Yonghao Zhuang, Joseph~E. Gonzalez, Ion Stoica, and Eric~P. Xing.
\newblock Vicuna: An open-source chatbot impressing gpt-4 with 90\%* chatgpt quality, March 2023.
\newblock URL \url{https://lmsys.org/blog/2023-03-30-vicuna/}.

\bibitem[Chollet(2019)]{chollet2019measureintelligence}
Fran{\c{c}}ois Chollet.
\newblock On the measure of intelligence.
\newblock \emph{CoRR}, abs/1911.01547, 2019.
\newblock URL \url{http://arxiv.org/abs/1911.01547}.

\bibitem[Chowdhery et~al.(2023)Chowdhery, Narang, Devlin, Bosma, Mishra, Roberts, Barham, Chung, Sutton, Gehrmann, Schuh, Shi, Tsvyashchenko, Maynez, Rao, Barnes, Tay, Shazeer, Prabhakaran, Reif, Du, Hutchinson, Pope, Bradbury, Austin, Isard, Gur{-}Ari, Yin, Duke, Levskaya, Ghemawat, Dev, Michalewski, Garcia, Misra, Robinson, Fedus, Zhou, Ippolito, Luan, Lim, Zoph, Spiridonov, Sepassi, Dohan, Agrawal, Omernick, Dai, Pillai, Pellat, Lewkowycz, Moreira, Child, Polozov, Lee, Zhou, Wang, Saeta, Diaz, Firat, Catasta, Wei, Meier{-}Hellstern, Eck, Dean, Petrov, and Fiedel]{chowdhery2023palm}
Aakanksha Chowdhery, Sharan Narang, Jacob Devlin, Maarten Bosma, Gaurav Mishra, Adam Roberts, Paul Barham, Hyung~Won Chung, Charles Sutton, Sebastian Gehrmann, Parker Schuh, Kensen Shi, Sasha Tsvyashchenko, Joshua Maynez, Abhishek Rao, Parker Barnes, Yi~Tay, Noam Shazeer, Vinodkumar Prabhakaran, Emily Reif, Nan Du, Ben Hutchinson, Reiner Pope, James Bradbury, Jacob Austin, Michael Isard, Guy Gur{-}Ari, Pengcheng Yin, Toju Duke, Anselm Levskaya, Sanjay Ghemawat, Sunipa Dev, Henryk Michalewski, Xavier Garcia, Vedant Misra, Kevin Robinson, Liam Fedus, Denny Zhou, Daphne Ippolito, David Luan, Hyeontaek Lim, Barret Zoph, Alexander Spiridonov, Ryan Sepassi, David Dohan, Shivani Agrawal, Mark Omernick, Andrew~M. Dai, Thanumalayan~Sankaranarayana Pillai, Marie Pellat, Aitor Lewkowycz, Erica Moreira, Rewon Child, Oleksandr Polozov, Katherine Lee, Zongwei Zhou, Xuezhi Wang, Brennan Saeta, Mark Diaz, Orhan Firat, Michele Catasta, Jason Wei, Kathy Meier{-}Hellstern, Douglas Eck, Jeff Dean, Slav Petrov, and Noah Fiedel.
\newblock Palm: Scaling language modeling with pathways.
\newblock \emph{J. Mach. Learn. Res.}, 24:\penalty0 240:1--240:113, 2023.
\newblock URL \url{https://jmlr.org/papers/v24/22-1144.html}.

\bibitem[DeepSeek{-}AI et~al.(2025)DeepSeek{-}AI, Guo, Yang, Zhang, Song, Zhang, Xu, Zhu, Ma, Wang, Bi, Zhang, Yu, Wu, Wu, Gou, Shao, Li, Gao, Liu, Xue, Wang, Wu, Feng, Lu, Zhao, Deng, Zhang, Ruan, Dai, Chen, Ji, Li, Lin, Dai, Luo, Hao, Chen, Li, Zhang, Bao, Xu, Wang, Ding, Xin, Gao, Qu, Li, Guo, Li, Wang, Chen, Yuan, Qiu, Li, Cai, Ni, Liang, Chen, Dong, Hu, Gao, Guan, Huang, Yu, Wang, Zhang, Zhao, Wang, Zhang, Xu, Xia, Zhang, Zhang, Tang, Li, Wang, Li, Tian, Huang, Zhang, Wang, Chen, Du, Ge, Zhang, Pan, Wang, Chen, Jin, Chen, Lu, Zhou, Chen, Ye, Wang, Yu, Zhou, Pan, and Li]{guo2025deepseek}
DeepSeek{-}AI, Daya Guo, Dejian Yang, Haowei Zhang, Junxiao Song, Ruoyu Zhang, Runxin Xu, Qihao Zhu, Shirong Ma, Peiyi Wang, Xiao Bi, Xiaokang Zhang, Xingkai Yu, Yu~Wu, Z.~F. Wu, Zhibin Gou, Zhihong Shao, Zhuoshu Li, Ziyi Gao, Aixin Liu, Bing Xue, Bingxuan Wang, Bochao Wu, Bei Feng, Chengda Lu, Chenggang Zhao, Chengqi Deng, Chenyu Zhang, Chong Ruan, Damai Dai, Deli Chen, Dongjie Ji, Erhang Li, Fangyun Lin, Fucong Dai, Fuli Luo, Guangbo Hao, Guanting Chen, Guowei Li, H.~Zhang, Han Bao, Hanwei Xu, Haocheng Wang, Honghui Ding, Huajian Xin, Huazuo Gao, Hui Qu, Hui Li, Jianzhong Guo, Jiashi Li, Jiawei Wang, Jingchang Chen, Jingyang Yuan, Junjie Qiu, Junlong Li, J.~L. Cai, Jiaqi Ni, Jian Liang, Jin Chen, Kai Dong, Kai Hu, Kaige Gao, Kang Guan, Kexin Huang, Kuai Yu, Lean Wang, Lecong Zhang, Liang Zhao, Litong Wang, Liyue Zhang, Lei Xu, Leyi Xia, Mingchuan Zhang, Minghua Zhang, Minghui Tang, Meng Li, Miaojun Wang, Mingming Li, Ning Tian, Panpan Huang, Peng Zhang, Qiancheng Wang, Qinyu Chen, Qiushi Du, Ruiqi Ge,
  Ruisong Zhang, Ruizhe Pan, Runji Wang, R.~J. Chen, R.~L. Jin, Ruyi Chen, Shanghao Lu, Shangyan Zhou, Shanhuang Chen, Shengfeng Ye, Shiyu Wang, Shuiping Yu, Shunfeng Zhou, Shuting Pan, and S.~S. Li.
\newblock Deepseek-r1: Incentivizing reasoning capability in llms via reinforcement learning.
\newblock \emph{CoRR}, abs/2501.12948, 2025.
\newblock \doi{10.48550/ARXIV.2501.12948}.
\newblock URL \url{https://doi.org/10.48550/arXiv.2501.12948}.

\bibitem[Dong et~al.(2024)Dong, Li, Dai, Zheng, Ma, Li, Xia, Xu, Wu, Chang, Sun, and Sui]{dong-etal-2024-survey}
Qingxiu Dong, Lei Li, Damai Dai, Ce~Zheng, Jingyuan Ma, Rui Li, Heming Xia, Jingjing Xu, Zhiyong Wu, Baobao Chang, Xu~Sun, and Zhifang Sui.
\newblock A survey on in-context learning.
\newblock In Yaser Al{-}Onaizan, Mohit Bansal, and Yun{-}Nung Chen (eds.), \emph{Proceedings of the 2024 Conference on Empirical Methods in Natural Language Processing, {EMNLP} 2024, Miami, FL, USA, November 12-16, 2024}, pp.\  1107--1128. Association for Computational Linguistics, 2024.
\newblock URL \url{https://aclanthology.org/2024.emnlp-main.64}.

\bibitem[Dubey et~al.(2024)Dubey, Jauhri, Pandey, Kadian, Al{-}Dahle, Letman, Mathur, Schelten, Yang, Fan, Goyal, Hartshorn, Yang, Mitra, Sravankumar, Korenev, Hinsvark, Rao, Zhang, Rodriguez, Gregerson, Spataru, Rozi{\`{e}}re, Biron, Tang, Chern, Caucheteux, Nayak, Bi, Marra, McConnell, Keller, Touret, Wu, Wong, Ferrer, Nikolaidis, Allonsius, Song, Pintz, Livshits, Esiobu, Choudhary, Mahajan, Garcia{-}Olano, Perino, Hupkes, Lakomkin, AlBadawy, Lobanova, Dinan, Smith, Radenovic, Zhang, Synnaeve, Lee, Anderson, Nail, Mialon, Pang, Cucurell, Nguyen, Korevaar, Xu, Touvron, Zarov, Ibarra, Kloumann, Misra, Evtimov, Copet, Lee, Geffert, Vranes, Park, Mahadeokar, Shah, van~der Linde, Billock, Hong, Lee, Fu, Chi, Huang, Liu, Wang, Yu, Bitton, Spisak, Park, Rocca, Johnstun, Saxe, Jia, Alwala, Upasani, Plawiak, Li, Heafield, Stone, and et~al.]{grattafiori2024llama3herdmodels}
Abhimanyu Dubey, Abhinav Jauhri, Abhinav Pandey, Abhishek Kadian, Ahmad Al{-}Dahle, Aiesha Letman, Akhil Mathur, Alan Schelten, Amy Yang, Angela Fan, Anirudh Goyal, Anthony Hartshorn, Aobo Yang, Archi Mitra, Archie Sravankumar, Artem Korenev, Arthur Hinsvark, Arun Rao, Aston Zhang, Aur{\'{e}}lien Rodriguez, Austen Gregerson, Ava Spataru, Baptiste Rozi{\`{e}}re, Bethany Biron, Binh Tang, Bobbie Chern, Charlotte Caucheteux, Chaya Nayak, Chloe Bi, Chris Marra, Chris McConnell, Christian Keller, Christophe Touret, Chunyang Wu, Corinne Wong, Cristian~Canton Ferrer, Cyrus Nikolaidis, Damien Allonsius, Daniel Song, Danielle Pintz, Danny Livshits, David Esiobu, Dhruv Choudhary, Dhruv Mahajan, Diego Garcia{-}Olano, Diego Perino, Dieuwke Hupkes, Egor Lakomkin, Ehab AlBadawy, Elina Lobanova, Emily Dinan, Eric~Michael Smith, Filip Radenovic, Frank Zhang, Gabriel Synnaeve, Gabrielle Lee, Georgia~Lewis Anderson, Graeme Nail, Gr{\'{e}}goire Mialon, Guan Pang, Guillem Cucurell, Hailey Nguyen, Hannah Korevaar, Hu~Xu, Hugo
  Touvron, Iliyan Zarov, Imanol~Arrieta Ibarra, Isabel~M. Kloumann, Ishan Misra, Ivan Evtimov, Jade Copet, Jaewon Lee, Jan Geffert, Jana Vranes, Jason Park, Jay Mahadeokar, Jeet Shah, Jelmer van~der Linde, Jennifer Billock, Jenny Hong, Jenya Lee, Jeremy Fu, Jianfeng Chi, Jianyu Huang, Jiawen Liu, Jie Wang, Jiecao Yu, Joanna Bitton, Joe Spisak, Jongsoo Park, Joseph Rocca, Joshua Johnstun, Joshua Saxe, Junteng Jia, Kalyan~Vasuden Alwala, Kartikeya Upasani, Kate Plawiak, Ke~Li, Kenneth Heafield, Kevin Stone, and et~al.
\newblock The llama 3 herd of models.
\newblock \emph{CoRR}, abs/2407.21783, 2024.
\newblock \doi{10.48550/ARXIV.2407.21783}.
\newblock URL \url{https://doi.org/10.48550/arXiv.2407.21783}.

\bibitem[Ethayarajh et~al.(2024)Ethayarajh, Xu, Muennighoff, Jurafsky, and Kiela]{ethayarajh2024ktomodelalignmentprospect}
Kawin Ethayarajh, Winnie Xu, Niklas Muennighoff, Dan Jurafsky, and Douwe Kiela.
\newblock {KTO:} model alignment as prospect theoretic optimization.
\newblock \emph{CoRR}, abs/2402.01306, 2024.
\newblock \doi{10.48550/ARXIV.2402.01306}.
\newblock URL \url{https://doi.org/10.48550/arXiv.2402.01306}.

\bibitem[Gao et~al.(2025)Gao, Sinha, and Das]{gao2025learning}
Xiang Gao, Ankita Sinha, and Kamalika Das.
\newblock Learning to search effective example sequences for in-context learning.
\newblock \emph{arXiv preprint arXiv:2503.08030}, 2025.

\bibitem[Gendron et~al.(2024)Gendron, Bao, Witbrock, and Dobbie]{gendron2024largelanguagemodelsstrong}
Ga{\"{e}}l Gendron, Qiming Bao, Michael Witbrock, and Gillian Dobbie.
\newblock Large language models are not strong abstract reasoners.
\newblock In \emph{Proceedings of the Thirty-Third International Joint Conference on Artificial Intelligence, {IJCAI} 2024, Jeju, South Korea, August 3-9, 2024}, pp.\  6270--6278. ijcai.org, 2024.
\newblock URL \url{https://www.ijcai.org/proceedings/2024/693}.

\bibitem[Hinton et~al.(2015)Hinton, Vinyals, and Dean]{hinton2015distillingknowledgeneuralnetwork}
Geoffrey~E. Hinton, Oriol Vinyals, and Jeffrey Dean.
\newblock Distilling the knowledge in a neural network.
\newblock \emph{CoRR}, abs/1503.02531, 2015.
\newblock URL \url{http://arxiv.org/abs/1503.02531}.

\bibitem[Hong et~al.(2024)Hong, Lee, and Thorne]{hong2024orpomonolithicpreferenceoptimization}
Jiwoo Hong, Noah Lee, and James Thorne.
\newblock {ORPO:} monolithic preference optimization without reference model.
\newblock In Yaser Al{-}Onaizan, Mohit Bansal, and Yun{-}Nung Chen (eds.), \emph{Proceedings of the 2024 Conference on Empirical Methods in Natural Language Processing, {EMNLP} 2024, Miami, FL, USA, November 12-16, 2024}, pp.\  11170--11189. Association for Computational Linguistics, 2024.
\newblock URL \url{https://aclanthology.org/2024.emnlp-main.626}.

\bibitem[Huang et~al.(2022)Huang, Chen, Yu, and McKeown]{huang2022incontextlearningdistillationtransferring}
Yukun Huang, Yanda Chen, Zhou Yu, and Kathleen~R. McKeown.
\newblock In-context learning distillation: Transferring few-shot learning ability of pre-trained language models.
\newblock \emph{CoRR}, abs/2212.10670, 2022.
\newblock \doi{10.48550/ARXIV.2212.10670}.
\newblock URL \url{https://doi.org/10.48550/arXiv.2212.10670}.

\bibitem[Hurst et~al.(2024)Hurst, Lerer, Goucher, Perelman, Ramesh, Clark, Ostrow, Welihinda, Hayes, Radford, Madry, Baker{-}Whitcomb, Beutel, Borzunov, Carney, Chow, Kirillov, Nichol, Paino, Renzin, Passos, Kirillov, Christakis, Conneau, Kamali, Jabri, Moyer, Tam, Crookes, Tootoonchian, Kumar, Vallone, Karpathy, Braunstein, Cann, Codispoti, Galu, Kondrich, Tulloch, Mishchenko, Baek, Jiang, Pelisse, Woodford, Gosalia, Dhar, Pantuliano, Nayak, Oliver, Zoph, Ghorbani, Leimberger, Rossen, Sokolowsky, Wang, Zweig, Hoover, Samic, McGrew, Spero, Giertler, Cheng, Lightcap, Walkin, Quinn, Guarraci, Hsu, Kellogg, Eastman, Lugaresi, Wainwright, Bassin, Hudson, Chu, Nelson, Li, Shern, Conger, Barette, Voss, Ding, Lu, Zhang, Beaumont, Hallacy, Koch, Gibson, Kim, Choi, McLeavey, Hesse, Fischer, Winter, Czarnecki, Jarvis, Wei, Koumouzelis, and Sherburn]{openai2024gpt4ocard}
Aaron Hurst, Adam Lerer, Adam~P. Goucher, Adam Perelman, Aditya Ramesh, Aidan Clark, AJ~Ostrow, Akila Welihinda, Alan Hayes, Alec Radford, Aleksander Madry, Alex Baker{-}Whitcomb, Alex Beutel, Alex Borzunov, Alex Carney, Alex Chow, Alex Kirillov, Alex Nichol, Alex Paino, Alex Renzin, Alex~Tachard Passos, Alexander Kirillov, Alexi Christakis, Alexis Conneau, Ali Kamali, Allan Jabri, Allison Moyer, Allison Tam, Amadou Crookes, Amin Tootoonchian, Ananya Kumar, Andrea Vallone, Andrej Karpathy, Andrew Braunstein, Andrew Cann, Andrew Codispoti, Andrew Galu, Andrew Kondrich, Andrew Tulloch, Andrey Mishchenko, Angela Baek, Angela Jiang, Antoine Pelisse, Antonia Woodford, Anuj Gosalia, Arka Dhar, Ashley Pantuliano, Avi Nayak, Avital Oliver, Barret Zoph, Behrooz Ghorbani, Ben Leimberger, Ben Rossen, Ben Sokolowsky, Ben Wang, Benjamin Zweig, Beth Hoover, Blake Samic, Bob McGrew, Bobby Spero, Bogo Giertler, Bowen Cheng, Brad Lightcap, Brandon Walkin, Brendan Quinn, Brian Guarraci, Brian Hsu, Bright Kellogg, Brydon
  Eastman, Camillo Lugaresi, Carroll~L. Wainwright, Cary Bassin, Cary Hudson, Casey Chu, Chad Nelson, Chak Li, Chan~Jun Shern, Channing Conger, Charlotte Barette, Chelsea Voss, Chen Ding, Cheng Lu, Chong Zhang, Chris Beaumont, Chris Hallacy, Chris Koch, Christian Gibson, Christina Kim, Christine Choi, Christine McLeavey, Christopher Hesse, Claudia Fischer, Clemens Winter, Coley Czarnecki, Colin Jarvis, Colin Wei, Constantin Koumouzelis, and Dane Sherburn.
\newblock Gpt-4o system card.
\newblock \emph{CoRR}, abs/2410.21276, 2024.
\newblock \doi{10.48550/ARXIV.2410.21276}.
\newblock URL \url{https://doi.org/10.48550/arXiv.2410.21276}.

\bibitem[Jaech et~al.(2024)Jaech, Kalai, Lerer, Richardson, El{-}Kishky, Low, Helyar, Madry, Beutel, Carney, Iftimie, Karpenko, Passos, Neitz, Prokofiev, Wei, Tam, Bennett, Kumar, Saraiva, Vallone, Duberstein, Kondrich, Mishchenko, Applebaum, Jiang, Nair, Zoph, Ghorbani, Rossen, Sokolowsky, Barak, McGrew, Minaiev, Hao, Baker, Houghton, McKinzie, Eastman, Lugaresi, Bassin, Hudson, Li, de~Bourcy, Voss, Shen, Zhang, Koch, Orsinger, Hesse, Fischer, Chan, Roberts, Kappler, Levy, Selsam, Dohan, Farhi, Mely, Robinson, Tsipras, Li, Oprica, Freeman, Zhang, Wong, Proehl, Cheung, Mitchell, Wallace, Ritter, Mays, Wang, Such, Raso, Leoni, Tsimpourlas, Song, von Lohmann, Sulit, Salmon, Parascandolo, Chabot, Zhao, Brockman, Leclerc, Salman, Bao, Sheng, Andrin, Bagherinezhad, Ren, Lightman, Chung, Kivlichan, O'Connell, Osband, Gilaberte, and Akkaya]{openai2024openaio1card}
Aaron Jaech, Adam Kalai, Adam Lerer, Adam Richardson, Ahmed El{-}Kishky, Aiden Low, Alec Helyar, Aleksander Madry, Alex Beutel, Alex Carney, Alex Iftimie, Alex Karpenko, Alex~Tachard Passos, Alexander Neitz, Alexander Prokofiev, Alexander Wei, Allison Tam, Ally Bennett, Ananya Kumar, Andre Saraiva, Andrea Vallone, Andrew Duberstein, Andrew Kondrich, Andrey Mishchenko, Andy Applebaum, Angela Jiang, Ashvin Nair, Barret Zoph, Behrooz Ghorbani, Ben Rossen, Benjamin Sokolowsky, Boaz Barak, Bob McGrew, Borys Minaiev, Botao Hao, Bowen Baker, Brandon Houghton, Brandon McKinzie, Brydon Eastman, Camillo Lugaresi, Cary Bassin, Cary Hudson, Chak~Ming Li, Charles de~Bourcy, Chelsea Voss, Chen Shen, Chong Zhang, Chris Koch, Chris Orsinger, Christopher Hesse, Claudia Fischer, Clive Chan, Dan Roberts, Daniel Kappler, Daniel Levy, Daniel Selsam, David Dohan, David Farhi, David Mely, David Robinson, Dimitris Tsipras, Doug Li, Dragos Oprica, Eben Freeman, Eddie Zhang, Edmund Wong, Elizabeth Proehl, Enoch Cheung, Eric Mitchell,
  Eric Wallace, Erik Ritter, Evan Mays, Fan Wang, Felipe~Petroski Such, Filippo Raso, Florencia Leoni, Foivos Tsimpourlas, Francis Song, Fred von Lohmann, Freddie Sulit, Geoff Salmon, Giambattista Parascandolo, Gildas Chabot, Grace Zhao, Greg Brockman, Guillaume Leclerc, Hadi Salman, Haiming Bao, Hao Sheng, Hart Andrin, Hessam Bagherinezhad, Hongyu Ren, Hunter Lightman, Hyung~Won Chung, Ian Kivlichan, Ian O'Connell, Ian Osband, Ignasi~Clavera Gilaberte, and Ilge Akkaya.
\newblock Openai o1 system card.
\newblock \emph{CoRR}, abs/2412.16720, 2024.
\newblock \doi{10.48550/ARXIV.2412.16720}.
\newblock URL \url{https://doi.org/10.48550/arXiv.2412.16720}.

\bibitem[Jiang et~al.(2023)Jiang, Sablayrolles, Mensch, Bamford, Chaplot, de~Las~Casas, Bressand, Lengyel, Lample, Saulnier, Lavaud, Lachaux, Stock, Scao, Lavril, Wang, Lacroix, and Sayed]{jiang2023mistral7b}
Albert~Q. Jiang, Alexandre Sablayrolles, Arthur Mensch, Chris Bamford, Devendra~Singh Chaplot, Diego de~Las~Casas, Florian Bressand, Gianna Lengyel, Guillaume Lample, Lucile Saulnier, L{\'{e}}lio~Renard Lavaud, Marie{-}Anne Lachaux, Pierre Stock, Teven~Le Scao, Thibaut Lavril, Thomas Wang, Timoth{\'{e}}e Lacroix, and William~El Sayed.
\newblock Mistral 7b.
\newblock \emph{CoRR}, abs/2310.06825, 2023.
\newblock \doi{10.48550/ARXIV.2310.06825}.
\newblock URL \url{https://doi.org/10.48550/arXiv.2310.06825}.

\bibitem[Jung et~al.(2024)Jung, West, Jiang, Brahman, Lu, Fisher, Sorensen, and Choi]{jung-etal-2024-impossible}
Jaehun Jung, Peter West, Liwei Jiang, Faeze Brahman, Ximing Lu, Jillian Fisher, Taylor Sorensen, and Yejin Choi.
\newblock Impossible distillation for paraphrasing and summarization: How to make high-quality lemonade out of small, low-quality model.
\newblock In Kevin Duh, Helena G{\'{o}}mez{-}Adorno, and Steven Bethard (eds.), \emph{Proceedings of the 2024 Conference of the North American Chapter of the Association for Computational Linguistics: Human Language Technologies (Volume 1: Long Papers), {NAACL} 2024, Mexico City, Mexico, June 16-21, 2024}, pp.\  4439--4454. Association for Computational Linguistics, 2024.
\newblock \doi{10.18653/V1/2024.NAACL-LONG.250}.
\newblock URL \url{https://doi.org/10.18653/v1/2024.naacl-long.250}.

\bibitem[Lake et~al.(2019)Lake, Linzen, and Baroni]{lake2019humanfewshotlearningcompositional}
Brenden~M. Lake, Tal Linzen, and Marco Baroni.
\newblock Human few-shot learning of compositional instructions.
\newblock In Ashok~K. Goel, Colleen~M. Seifert, and Christian Freksa (eds.), \emph{Proceedings of the 41th Annual Meeting of the Cognitive Science Society, CogSci 2019: Creativity + Cognition + Computation, Montreal, Canada, July 24-27, 2019}, pp.\  611--617. cognitivesciencesociety.org, 2019.
\newblock URL \url{https://mindmodeling.org/cogsci2019/papers/0123/index.html}.

\bibitem[Levy et~al.(2023)Levy, Bogin, and Berant]{levy-etal-2023-diverse}
Itay Levy, Ben Bogin, and Jonathan Berant.
\newblock Diverse demonstrations improve in-context compositional generalization.
\newblock In Anna Rogers, Jordan~L. Boyd{-}Graber, and Naoaki Okazaki (eds.), \emph{Proceedings of the 61st Annual Meeting of the Association for Computational Linguistics (Volume 1: Long Papers), {ACL} 2023, Toronto, Canada, July 9-14, 2023}, pp.\  1401--1422. Association for Computational Linguistics, 2023.
\newblock \doi{10.18653/V1/2023.ACL-LONG.78}.
\newblock URL \url{https://doi.org/10.18653/v1/2023.acl-long.78}.

\bibitem[Li et~al.(2024{\natexlab{a}})Li, Cao, Jin, Chen, Liu, and Zhao]{li2025mirageevaluatingexplaininginductive}
Jiachun Li, Pengfei Cao, Zhuoran Jin, Yubo Chen, Kang Liu, and Jun Zhao.
\newblock {MIRAGE:} evaluating and explaining inductive reasoning process in language models.
\newblock \emph{CoRR}, abs/2410.09542, 2024{\natexlab{a}}.
\newblock \doi{10.48550/ARXIV.2410.09542}.
\newblock URL \url{https://doi.org/10.48550/arXiv.2410.09542}.

\bibitem[Li et~al.(2022)Li, Chen, Shen, Chen, Zhang, Li, Wang, Qian, Peng, Mao, Chen, and Yan]{li2022explanationslargelanguagemodels}
Shiyang Li, Jianshu Chen, Yelong Shen, Zhiyu Chen, Xinlu Zhang, Zekun Li, Hong Wang, Jing Qian, Baolin Peng, Yi~Mao, Wenhu Chen, and Xifeng Yan.
\newblock Explanations from large language models make small reasoners better.
\newblock \emph{CoRR}, abs/2210.06726, 2022.
\newblock \doi{10.48550/ARXIV.2210.06726}.
\newblock URL \url{https://doi.org/10.48550/arXiv.2210.06726}.

\bibitem[Li et~al.(2024{\natexlab{b}})Li, He, Wu, Yang, Xu, jun Jun, Liu, Liu, and Zhao]{li-etal-2024-mode}
Xiang Li, Shizhu He, Jiayu Wu, Zhao Yang, Yao Xu, Yang jun Jun, Haifeng Liu, Kang Liu, and Jun Zhao.
\newblock Mode-cotd: Chain-of-thought distillation for complex reasoning tasks with mixture of decoupled lora-experts.
\newblock In Nicoletta Calzolari, Min{-}Yen Kan, V{\'{e}}ronique Hoste, Alessandro Lenci, Sakriani Sakti, and Nianwen Xue (eds.), \emph{Proceedings of the 2024 Joint International Conference on Computational Linguistics, Language Resources and Evaluation, {LREC/COLING} 2024, 20-25 May, 2024, Torino, Italy}, pp.\  11475--11485. {ELRA} and {ICCL}, 2024{\natexlab{b}}.
\newblock URL \url{https://aclanthology.org/2024.lrec-main.1003}.

\bibitem[Li et~al.(2023{\natexlab{a}})Li, Lv, Yan, Lin, Zhu, Ni, Xie, Wang, and Qiu]{li-etal-2023-unified}
Xiaonan Li, Kai Lv, Hang Yan, Tianyang Lin, Wei Zhu, Yuan Ni, Guotong Xie, Xiaoling Wang, and Xipeng Qiu.
\newblock Unified demonstration retriever for in-context learning.
\newblock In Anna Rogers, Jordan~L. Boyd{-}Graber, and Naoaki Okazaki (eds.), \emph{Proceedings of the 61st Annual Meeting of the Association for Computational Linguistics (Volume 1: Long Papers), {ACL} 2023, Toronto, Canada, July 9-14, 2023}, pp.\  4644--4668. Association for Computational Linguistics, 2023{\natexlab{a}}.
\newblock \doi{10.18653/V1/2023.ACL-LONG.256}.
\newblock URL \url{https://doi.org/10.18653/v1/2023.acl-long.256}.

\bibitem[Li et~al.(2023{\natexlab{b}})Li, Ildiz, Papailiopoulos, and Oymak]{DBLP:conf/icml/LiIPO23}
Yingcong Li, Muhammed~Emrullah Ildiz, Dimitris Papailiopoulos, and Samet Oymak.
\newblock Transformers as algorithms: Generalization and stability in in-context learning.
\newblock In Andreas Krause, Emma Brunskill, Kyunghyun Cho, Barbara Engelhardt, Sivan Sabato, and Jonathan Scarlett (eds.), \emph{International Conference on Machine Learning, {ICML} 2023, 23-29 July 2023, Honolulu, Hawaii, {USA}}, volume 202 of \emph{Proceedings of Machine Learning Research}, pp.\  19565--19594. {PMLR}, 2023{\natexlab{b}}.
\newblock URL \url{https://proceedings.mlr.press/v202/li23l.html}.

\bibitem[Luo et~al.(2024)Luo, Xu, Liu, Pasupat, and Kazemi]{DBLP:journals/corr/abs-2401-11624}
Man Luo, Xin Xu, Yue Liu, Panupong Pasupat, and Mehran Kazemi.
\newblock In-context learning with retrieved demonstrations for language models: {A} survey.
\newblock \emph{CoRR}, abs/2401.11624, 2024.
\newblock \doi{10.48550/ARXIV.2401.11624}.
\newblock URL \url{https://doi.org/10.48550/arXiv.2401.11624}.

\bibitem[Min et~al.(2022{\natexlab{a}})Min, Lewis, Zettlemoyer, and Hajishirzi]{metaICL}
Sewon Min, Mike Lewis, Luke Zettlemoyer, and Hannaneh Hajishirzi.
\newblock Metaicl: Learning to learn in context.
\newblock In Marine Carpuat, Marie{-}Catherine de~Marneffe, and Iv{\'{a}}n Vladimir~Meza Ru{\'{\i}}z (eds.), \emph{Proceedings of the 2022 Conference of the North American Chapter of the Association for Computational Linguistics: Human Language Technologies, {NAACL} 2022, Seattle, WA, United States, July 10-15, 2022}, pp.\  2791--2809. Association for Computational Linguistics, 2022{\natexlab{a}}.
\newblock \doi{10.18653/V1/2022.NAACL-MAIN.201}.
\newblock URL \url{https://doi.org/10.18653/v1/2022.naacl-main.201}.

\bibitem[Min et~al.(2022{\natexlab{b}})Min, Lyu, Holtzman, Artetxe, Lewis, Hajishirzi, and Zettlemoyer]{min-etal-2022-rethinking}
Sewon Min, Xinxi Lyu, Ari Holtzman, Mikel Artetxe, Mike Lewis, Hannaneh Hajishirzi, and Luke Zettlemoyer.
\newblock Rethinking the role of demonstrations: What makes in-context learning work?
\newblock In Yoav Goldberg, Zornitsa Kozareva, and Yue Zhang (eds.), \emph{Proceedings of the 2022 Conference on Empirical Methods in Natural Language Processing, {EMNLP} 2022, Abu Dhabi, United Arab Emirates, December 7-11, 2022}, pp.\  11048--11064. Association for Computational Linguistics, 2022{\natexlab{b}}.
\newblock \doi{10.18653/V1/2022.EMNLP-MAIN.759}.
\newblock URL \url{https://doi.org/10.18653/v1/2022.emnlp-main.759}.

\bibitem[Mukherjee et~al.(2023)Mukherjee, Mitra, Jawahar, Agarwal, Palangi, and Awadallah]{mukherjee2023orcaprogressivelearningcomplex}
Subhabrata Mukherjee, Arindam Mitra, Ganesh Jawahar, Sahaj Agarwal, Hamid Palangi, and Ahmed Awadallah.
\newblock Orca: Progressive learning from complex explanation traces of {GPT-4}.
\newblock \emph{CoRR}, abs/2306.02707, 2023.
\newblock \doi{10.48550/ARXIV.2306.02707}.
\newblock URL \url{https://doi.org/10.48550/arXiv.2306.02707}.

\bibitem[OpenAI(2023)]{openai2024gpt4technicalreport}
OpenAI.
\newblock {GPT-4} technical report.
\newblock \emph{CoRR}, abs/2303.08774, 2023.
\newblock \doi{10.48550/ARXIV.2303.08774}.
\newblock URL \url{https://doi.org/10.48550/arXiv.2303.08774}.

\bibitem[Ouyang et~al.(2022)Ouyang, Wu, Jiang, Almeida, Wainwright, Mishkin, Zhang, Agarwal, Slama, Ray, Schulman, Hilton, Kelton, Miller, Simens, Askell, Welinder, Christiano, Leike, and Lowe]{ouyang2022training}
Long Ouyang, Jeffrey Wu, Xu~Jiang, Diogo Almeida, Carroll~L. Wainwright, Pamela Mishkin, Chong Zhang, Sandhini Agarwal, Katarina Slama, Alex Ray, John Schulman, Jacob Hilton, Fraser Kelton, Luke Miller, Maddie Simens, Amanda Askell, Peter Welinder, Paul~F. Christiano, Jan Leike, and Ryan Lowe.
\newblock Training language models to follow instructions with human feedback.
\newblock In Sanmi Koyejo, S.~Mohamed, A.~Agarwal, Danielle Belgrave, K.~Cho, and A.~Oh (eds.), \emph{Advances in Neural Information Processing Systems 35: Annual Conference on Neural Information Processing Systems 2022, NeurIPS 2022, New Orleans, LA, USA, November 28 - December 9, 2022}, 2022.
\newblock URL \url{http://papers.nips.cc/paper\_files/paper/2022/hash/b1efde53be364a73914f58805a001731-Abstract-Conference.html}.

\bibitem[Peirce(1868)]{peirce1868questions}
C.~S. Peirce.
\newblock Questions concerning certain faculties claimed for man.
\newblock \emph{The Journal of Speculative Philosophy}, 2\penalty0 (2):\penalty0 103--114, 1868.
\newblock ISSN 0891625X, 15279383.
\newblock URL \url{http://www.jstor.org/stable/25665643}.

\bibitem[Peng et~al.(2024)Peng, Ding, Yuan, Liu, Zhang, Ouyang, and Tao]{peng-etal-2024-revisiting}
Keqin Peng, Liang Ding, Yancheng Yuan, Xuebo Liu, Min Zhang, Yuanxin Ouyang, and Dacheng Tao.
\newblock Revisiting demonstration selection strategies in in-context learning.
\newblock In Lun{-}Wei Ku, Andre Martins, and Vivek Srikumar (eds.), \emph{Proceedings of the 62nd Annual Meeting of the Association for Computational Linguistics (Volume 1: Long Papers), {ACL} 2024, Bangkok, Thailand, August 11-16, 2024}, pp.\  9090--9101. Association for Computational Linguistics, 2024.
\newblock \doi{10.18653/V1/2024.ACL-LONG.492}.
\newblock URL \url{https://doi.org/10.18653/v1/2024.acl-long.492}.

\bibitem[Qiu et~al.(2024)Qiu, Jiang, Lu, Sclar, Pyatkin, Bhagavatula, Wang, Kim, Choi, Dziri, and Ren]{qiu2023phenomenal}
Linlu Qiu, Liwei Jiang, Ximing Lu, Melanie Sclar, Valentina Pyatkin, Chandra Bhagavatula, Bailin Wang, Yoon Kim, Yejin Choi, Nouha Dziri, and Xiang Ren.
\newblock Phenomenal yet puzzling: Testing inductive reasoning capabilities of language models with hypothesis refinement.
\newblock In \emph{The Twelfth International Conference on Learning Representations, {ICLR} 2024, Vienna, Austria, May 7-11, 2024}. OpenReview.net, 2024.
\newblock URL \url{https://openreview.net/forum?id=bNt7oajl2a}.

\bibitem[Rafailov et~al.(2023)Rafailov, Sharma, Mitchell, Manning, Ermon, and Finn]{rafailov2024directpreferenceoptimizationlanguage}
Rafael Rafailov, Archit Sharma, Eric Mitchell, Christopher~D. Manning, Stefano Ermon, and Chelsea Finn.
\newblock Direct preference optimization: Your language model is secretly a reward model.
\newblock In Alice Oh, Tristan Naumann, Amir Globerson, Kate Saenko, Moritz Hardt, and Sergey Levine (eds.), \emph{Advances in Neural Information Processing Systems 36: Annual Conference on Neural Information Processing Systems 2023, NeurIPS 2023, New Orleans, LA, USA, December 10 - 16, 2023}, 2023.
\newblock URL \url{http://papers.nips.cc/paper\_files/paper/2023/hash/a85b405ed65c6477a4fe8302b5e06ce7-Abstract-Conference.html}.

\bibitem[Romero et~al.(2015)Romero, Ballas, Kahou, Chassang, Gatta, and Bengio]{romero2015fitnetshintsdeepnets}
Adriana Romero, Nicolas Ballas, Samira~Ebrahimi Kahou, Antoine Chassang, Carlo Gatta, and Yoshua Bengio.
\newblock Fitnets: Hints for thin deep nets, 2015.
\newblock URL \url{https://arxiv.org/abs/1412.6550}.

\bibitem[Rule et~al.(2020)Rule, Tenenbaum, and Piantadosi]{rule2020child}
Joshua~S Rule, Joshua~B Tenenbaum, and Steven~T Piantadosi.
\newblock The child as hacker.
\newblock \emph{Trends in cognitive sciences}, 24\penalty0 (11):\penalty0 900--915, 2020.

\bibitem[Shao et~al.(2025)Shao, Li, Ma, Li, Song, Cheng, Li, Li, Wang, Guo, Yan, Qiu, Huang, and Lin]{shao-etal-2025-case2code}
Yunfan Shao, Linyang Li, Yichuan Ma, Peiji Li, Demin Song, Qinyuan Cheng, Shimin Li, Xiaonan Li, Pengyu Wang, Qipeng Guo, Hang Yan, Xipeng Qiu, Xuanjing Huang, and Dahua Lin.
\newblock Case2code: Scalable synthetic data for code generation.
\newblock In Owen Rambow, Leo Wanner, Marianna Apidianaki, Hend Al{-}Khalifa, Barbara~Di Eugenio, and Steven Schockaert (eds.), \emph{Proceedings of the 31st International Conference on Computational Linguistics, {COLING} 2025, Abu Dhabi, UAE, January 19-24, 2025}, pp.\  11056--11069. Association for Computational Linguistics, 2025.
\newblock URL \url{https://aclanthology.org/2025.coling-main.733/}.

\bibitem[Taori et~al.(2023)Taori, Gulrajani, Zhang, Dubois, Li, Guestrin, Liang, and Hashimoto]{alpaca}
Rohan Taori, Ishaan Gulrajani, Tianyi Zhang, Yann Dubois, Xuechen Li, Carlos Guestrin, Percy Liang, and Tatsunori~B. Hashimoto.
\newblock Stanford alpaca: An instruction-following llama model.
\newblock \url{https://github.com/tatsu-lab/stanford_alpaca}, 2023.

\bibitem[Wang et~al.(2024)Wang, Zelikman, Poesia, Pu, Haber, and Goodman]{wang2023hypothesis}
Ruocheng Wang, Eric Zelikman, Gabriel Poesia, Yewen Pu, Nick Haber, and Noah~D. Goodman.
\newblock Hypothesis search: Inductive reasoning with language models.
\newblock In \emph{The Twelfth International Conference on Learning Representations, {ICLR} 2024, Vienna, Austria, May 7-11, 2024}. OpenReview.net, 2024.
\newblock URL \url{https://openreview.net/forum?id=G7UtIGQmjm}.

\bibitem[Wei et~al.(2023{\natexlab{a}})Wei, Hou, Lampinen, Chen, Huang, Tay, Chen, Lu, Zhou, Ma, and Le]{wei-etal-2023-symbol}
Jerry~W. Wei, Le~Hou, Andrew~K. Lampinen, Xiangning Chen, Da~Huang, Yi~Tay, Xinyun Chen, Yifeng Lu, Denny Zhou, Tengyu Ma, and Quoc~V. Le.
\newblock Symbol tuning improves in-context learning in language models.
\newblock In Houda Bouamor, Juan Pino, and Kalika Bali (eds.), \emph{Proceedings of the 2023 Conference on Empirical Methods in Natural Language Processing, {EMNLP} 2023, Singapore, December 6-10, 2023}, pp.\  968--979. Association for Computational Linguistics, 2023{\natexlab{a}}.
\newblock \doi{10.18653/V1/2023.EMNLP-MAIN.61}.
\newblock URL \url{https://doi.org/10.18653/v1/2023.emnlp-main.61}.

\bibitem[Wei et~al.(2023{\natexlab{b}})Wei, Wei, Tay, Tran, Webson, Lu, Chen, Liu, Huang, Zhou, and Ma]{wei2023larger}
Jerry~W. Wei, Jason Wei, Yi~Tay, Dustin Tran, Albert Webson, Yifeng Lu, Xinyun Chen, Hanxiao Liu, Da~Huang, Denny Zhou, and Tengyu Ma.
\newblock Larger language models do in-context learning differently.
\newblock \emph{CoRR}, abs/2303.03846, 2023{\natexlab{b}}.
\newblock \doi{10.48550/ARXIV.2303.03846}.
\newblock URL \url{https://doi.org/10.48550/arXiv.2303.03846}.

\bibitem[West et~al.(2022)West, Bhagavatula, Hessel, Hwang, Jiang, Bras, Lu, Welleck, and Choi]{west-etal-2022-symbolic}
Peter West, Chandra Bhagavatula, Jack Hessel, Jena~D. Hwang, Liwei Jiang, Ronan~Le Bras, Ximing Lu, Sean Welleck, and Yejin Choi.
\newblock Symbolic knowledge distillation: from general language models to commonsense models.
\newblock In Marine Carpuat, Marie{-}Catherine de~Marneffe, and Iv{\'{a}}n Vladimir~Meza Ru{\'{\i}}z (eds.), \emph{Proceedings of the 2022 Conference of the North American Chapter of the Association for Computational Linguistics: Human Language Technologies, {NAACL} 2022, Seattle, WA, United States, July 10-15, 2022}, pp.\  4602--4625. Association for Computational Linguistics, 2022.
\newblock \doi{10.18653/V1/2022.NAACL-MAIN.341}.
\newblock URL \url{https://doi.org/10.18653/v1/2022.naacl-main.341}.

\bibitem[Xie et~al.(2024)Xie, Majumder, Zhao, Maeda, Yamada, Wakaki, and McAuley]{xie-etal-2024-shot-dialogue}
Zhouhang Xie, Bodhisattwa~Prasad Majumder, Mengjie Zhao, Yoshinori Maeda, Keiichi Yamada, Hiromi Wakaki, and Julian~J. McAuley.
\newblock Few-shot dialogue strategy learning for motivational interviewing via inductive reasoning.
\newblock In Lun{-}Wei Ku, Andre Martins, and Vivek Srikumar (eds.), \emph{Findings of the Association for Computational Linguistics, {ACL} 2024, Bangkok, Thailand and virtual meeting, August 11-16, 2024}, pp.\  13207--13219. Association for Computational Linguistics, 2024.
\newblock \doi{10.18653/V1/2024.FINDINGS-ACL.782}.
\newblock URL \url{https://doi.org/10.18653/v1/2024.findings-acl.782}.

\bibitem[Xu et~al.(2023)Xu, Zhu, Wang, and Zhang]{xu-etal-2023-unleash}
Xin Xu, Yuqi Zhu, Xiaohan Wang, and Ningyu Zhang.
\newblock How to unleash the power of large language models for few-shot relation extraction?
\newblock In Nafise~Sadat Moosavi, Iryna Gurevych, Yufang Hou, Gyuwan Kim, Young~Jin Kim, Tal Schuster, and Ameeta Agrawal (eds.), \emph{Proceedings of The Fourth Workshop on Simple and Efficient Natural Language Processing, SustaiNLP 2023, Toronto, Canada (Hybrid), July 13, 2023}, pp.\  190--200. Association for Computational Linguistics, 2023.
\newblock \doi{10.18653/V1/2023.SUSTAINLP-1.13}.
\newblock URL \url{https://doi.org/10.18653/v1/2023.sustainlp-1.13}.

\bibitem[Xu et~al.(2024)Xu, Li, Vaezipoor, Sanner, and Khalil]{xu2024llmsabstractionreasoningcorpus}
Yudong Xu, Wenhao Li, Pashootan Vaezipoor, Scott Sanner, and Elias~Boutros Khalil.
\newblock Llms and the abstraction and reasoning corpus: Successes, failures, and the importance of object-based representations.
\newblock \emph{Trans. Mach. Learn. Res.}, 2024, 2024.
\newblock URL \url{https://openreview.net/forum?id=E8m8oySvPJ}.

\bibitem[Yang et~al.(2024)Yang, Dong, Du, Cheng, Cambria, Liu, Gao, and Wei]{yang-etal-2024-language}
Zonglin Yang, Li~Dong, Xinya Du, Hao Cheng, Erik Cambria, Xiaodong Liu, Jianfeng Gao, and Furu Wei.
\newblock Language models as inductive reasoners.
\newblock In Yvette Graham and Matthew Purver (eds.), \emph{Proceedings of the 18th Conference of the European Chapter of the Association for Computational Linguistics, {EACL} 2024 - Volume 1: Long Papers, St. Julian's, Malta, March 17-22, 2024}, pp.\  209--225. Association for Computational Linguistics, 2024.
\newblock URL \url{https://aclanthology.org/2024.eacl-long.13}.

\bibitem[Zhang et~al.(2021)Zhang, Jia, Edmonds, Zhu, and Zhu]{zhang2021acreabstractcausalreasoning}
Chi Zhang, Baoxiong Jia, Mark Edmonds, Song{-}Chun Zhu, and Yixin Zhu.
\newblock {ACRE:} abstract causal reasoning beyond covariation.
\newblock In \emph{{IEEE} Conference on Computer Vision and Pattern Recognition, {CVPR} 2021, virtual, June 19-25, 2021}, pp.\  10643--10653. Computer Vision Foundation / {IEEE}, 2021.
\newblock \doi{10.1109/CVPR46437.2021.01050}.
\newblock URL \url{https://openaccess.thecvf.com/content/CVPR2021/html/Zhang\_ACRE\_Abstract\_Causal\_REasoning\_Beyond\_Covariation\_CVPR\_2021\_paper.html}.

\bibitem[Zhang et~al.(2023)Zhang, Zhou, and Liu]{DBLP:conf/nips/ZhangZ023}
Yuanhan Zhang, Kaiyang Zhou, and Ziwei Liu.
\newblock What makes good examples for visual in-context learning?
\newblock In Alice Oh, Tristan Naumann, Amir Globerson, Kate Saenko, Moritz Hardt, and Sergey Levine (eds.), \emph{Advances in Neural Information Processing Systems 36: Annual Conference on Neural Information Processing Systems 2023, NeurIPS 2023, New Orleans, LA, USA, December 10 - 16, 2023}, 2023.
\newblock URL \url{http://papers.nips.cc/paper\_files/paper/2023/hash/398ae57ed4fda79d0781c65c926d667b-Abstract-Conference.html}.

\end{thebibliography}
\bibliographystyle{colm2025_conference}

\appendix
\section{Prompts}
\label{app:prompts}

\begin{table}
\resizebox{\textwidth}{!}{
\begin{tabular}{p{2cm}|p{4cm}|p{4cm}|p{4cm}}
\toprule
\textbf{Task} & \textbf{Generate Rule ($\phi_{rg}$)} & \textbf{Follow Rule ($\phi_{rf}$)} & \textbf{Direct Few-shot ($\phi$)} \\
\midrule
List Function & Consider the following input-output examples. Both the input and the output are Python lists. Based on the example shown below, identify the rule that can be used to convert the input into the output. Your output should solely contain the rule. & The following problem contains a rule and an input. Use the rule to produce the output from the given input. Your output should solely be a valid Python list that represents the output. No other text or description should be present. & Consider the following input-output examples. Both the input and the output are Python lists. Based on the example shown above, produce the output from the following input. No other text or description should be present. \\ \midrule
1D ARC & Consider the following input-output examples. Both the input and the output are Python lists. Based on the example shown below, identify the rule that can be used to convert the input into the output. Your output should solely contain the rule. & The following problem contains a rule and an input. Use the rule to produce the output from the given input. Your output should solely be a valid Python list that represents the output. No other text or description should be present. & Consider the following input-output examples. Both the input and the output are Python lists. Based on the example shown above, produce the output from the following input. No other text or description should be present. \\ \midrule
ACRE & Generate a rule that maps the following inputs to their corresponding outputs. Each example is an input-output pair. The input is a list of objects. The presence of certain objects will trigger the light to turn on. The output is either 'on', 'off', or 'undetermined', indicating the state of the light or if the state of the light cannot be determined. Your output should solely contain the rule. & Generate an output corresponding to the given input based on the following rule. Each example is an input-output pair. The input is a list of objects. The presence of certain objects will trigger the light to turn on. The output is either 'on', 'off', or 'undetermined', indicating the state of the light or if the state of the light cannot be determined. & Consider the following few-shot examples. Each example is an input-output pair. The input is a list of objects. The presence of certain objects will trigger the light to turn on. The output is either 'on', 'off', or 'undetermined', indicating the state of the light or if the state of the light cannot be determined. Based on the example shown above, produce the output from the following input. No other text or description should be present. \\ \midrule
MiniSCAN & Generate rules that map the following inputs to their corresponding outputs. Specify the priority of the rules if necessary. Try to make your rules as minimal and generally applicable as possible. & Generate an output corresponding to the given input based on the following rules. The output is a sequence of tokens joined by spaces. & Consider the following input-output few-shot examples. Based on the example shown above, produce the output from the following input. No other text or description should be present. \\ 
 \bottomrule
\end{tabular}
}
\caption{In-context learning prompts}
\label{tab:prompts}
\end{table}

\section{Decoding Strategy}
\label{app:temperature}

\begin{figure}[htbp]
    \centering

    \begin{subfigure}[b]{0.49\textwidth}
        \centering
        \includegraphics[width=\linewidth]{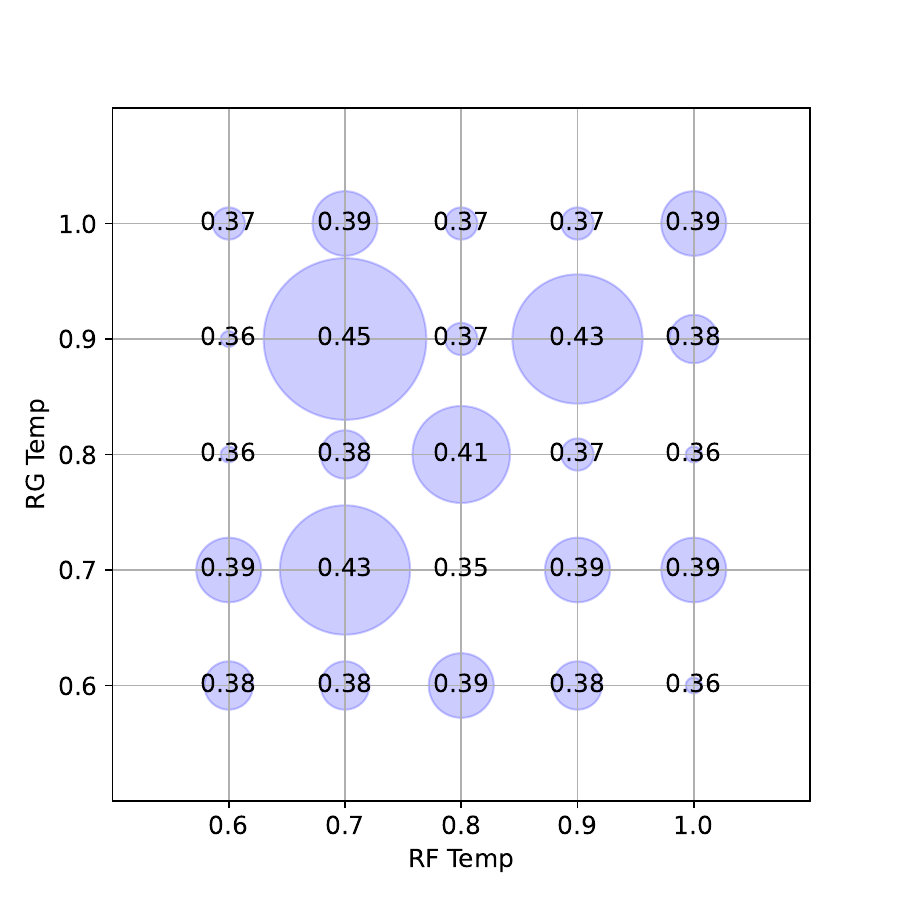}
        \caption{\proposed-Mistral}
    \end{subfigure}
    \hfill
    \begin{subfigure}[b]{0.49\textwidth}
        \centering
        \includegraphics[width=\linewidth]{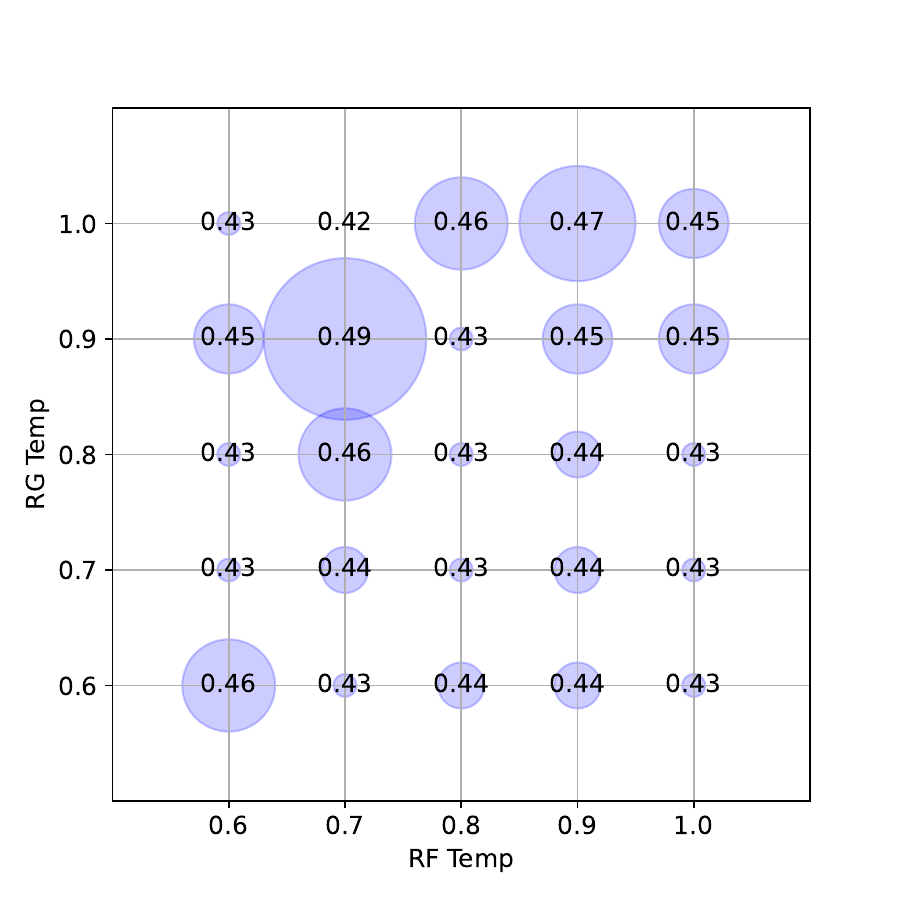}
        \caption{\proposed-Llama}
    \end{subfigure}

    \caption{The impact of decoding temperature on the performance of a) \proposed-Mistral b) \proposed-Llama on List function task. We perform grid search within the range between $0.6$ and $1.0$ for both rule generation and rule following and find the best performance for a rule generation temperature of $0.9$ and rule following temperature of $0.7$.}
    \label{fig:temperature_search}
\end{figure}

\section{Hypotheses Quality Analysis}
\label{app:data_quality}

\begin{figure}[htbp]
    \centering

    \begin{subfigure}[b]{0.49\textwidth}
        \centering
        \includegraphics[width=\linewidth]{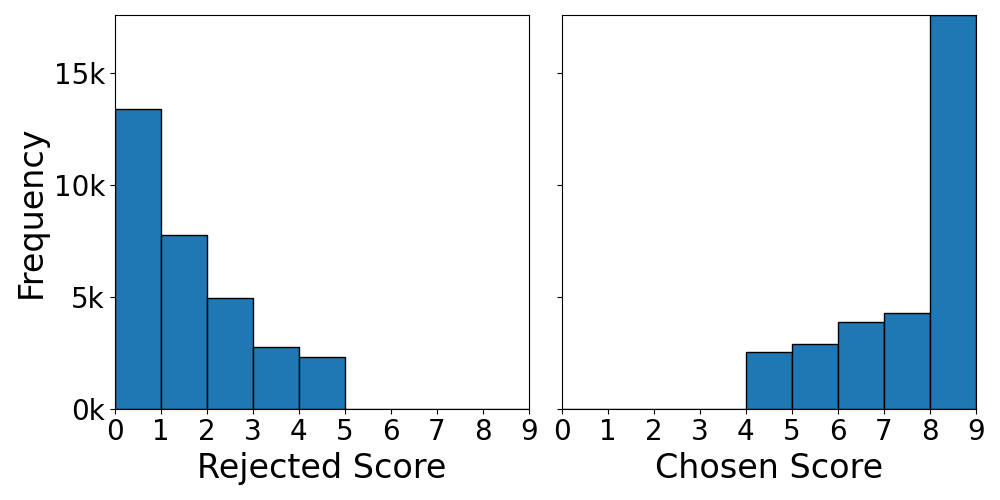}
        \caption{List Function (n=8,d=3)}
    \end{subfigure}
    \hfill
    \begin{subfigure}[b]{0.49\textwidth}
        \centering
        \includegraphics[width=\linewidth]{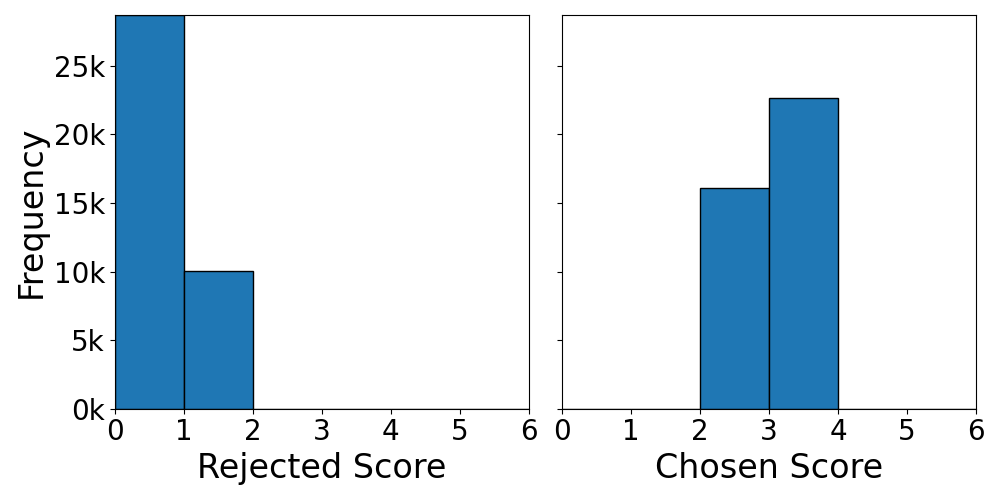}
        \caption{1D ARC (n=3,d=1)}
    \end{subfigure}

    \vskip\baselineskip
    \begin{subfigure}[b]{0.49\textwidth}
        \centering
        \includegraphics[width=\linewidth]{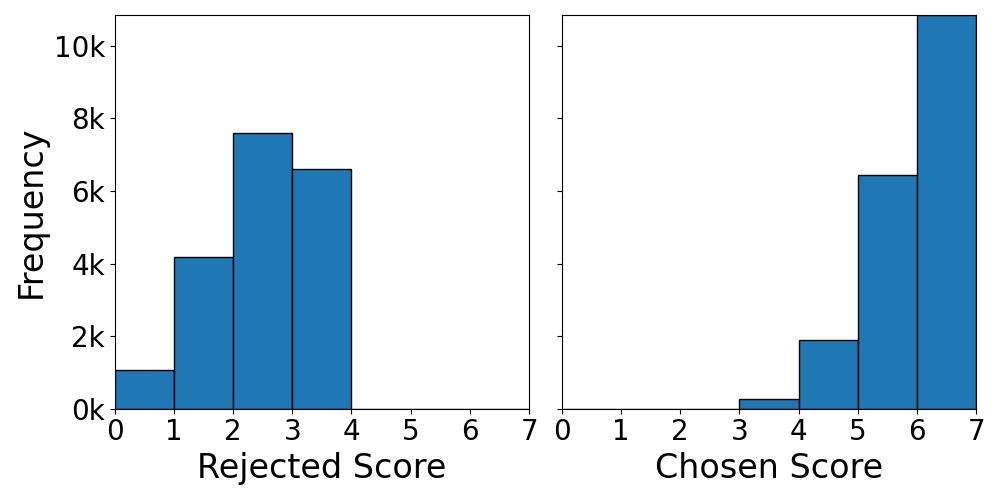}
        \caption{ACRE (n=6,d=2)}
    \end{subfigure}
    \hfill
    \begin{subfigure}[b]{0.49\textwidth}
        \centering
        \includegraphics[width=\linewidth]{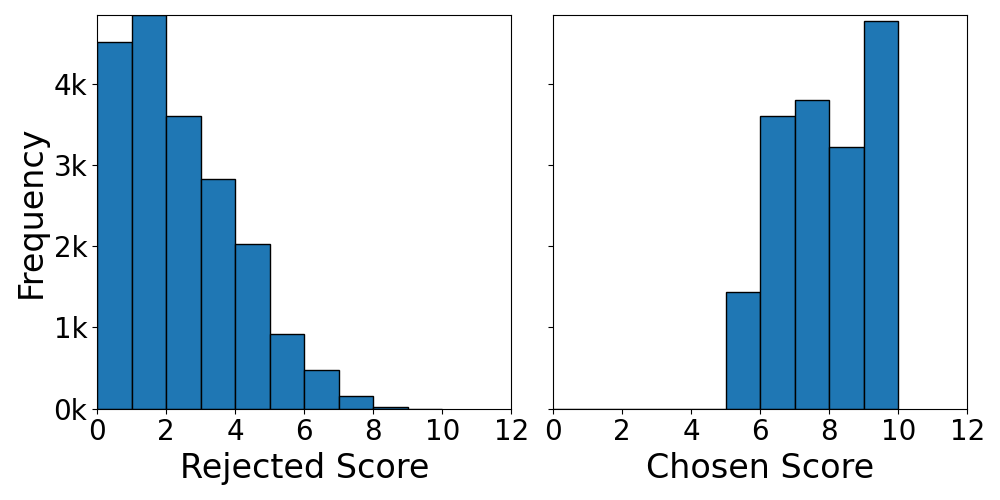}
        \caption{MiniSCAN (n=14,d=4)}
    \end{subfigure}

    \caption{Score distribution of chosen and rejected rules in the augmented dataset via noisy fitness estimation. $n$ denotes the maximum number of few-shot demonstrations, and $d$ denotes the minimum score difference between a (chosen, rejected) rule pair.}
    \label{fig:data_quality}
\end{figure}

\end{document}